\documentclass[10pt,twocolumn,letterpaper]{article}

\usepackage{cvpr}

\usepackage{graphicx}
\graphicspath{{figures/main/}{figures/supplement/}}

 \usepackage{multirow}
\usepackage{booktabs}   \usepackage{siunitx}    \usepackage{xcolor}
\usepackage{colortbl}
\usepackage{bbm,eucal}

\usepackage[most]{tcolorbox}
\tcbuselibrary{skins, breakable}

\definecolor{cvprblue}{rgb}{0.21,0.49,0.74}
\usepackage[pagebackref,breaklinks,colorlinks,allcolors=cvprblue]{hyperref}

\def\paperID{} \def\confName{CVPR}
\def\confYear{2026}

\title{Eliciting Complex Spatial Reasoning in MLLMs \\ through Wide-Baseline Matching}
\author{
Hao Zhong$^{1,2}$\thanks{Equal contributions. }\quad
Muzhi Zhu$^{1,2}$\footnotemark[1]\quad
Shenyan Zeng$^{1}$\footnotemark[1]\quad
Anzhou Li$^{2}$\quad
Cong Chen$^{1,2}$\\
Hua Geng$^{1}$\quad
Duochao Shi$^{1}$\quad
Wentao Ye$^{2}$\quad
Tao Lin$^{3,2}$\thanks{Corresponding authors. }\quad
Hao Chen$^{1}$\footnotemark[2]\quad
Chunhua Shen$^{1,2}$\footnotemark[2]\\[0.25cm]
\small 
$^{1}$ State Key Laboratory of CAD \& CG, Zhejiang University~~~~~
$^{2}$ Ant Group~~~~~
$^{3}$ Westlake University
}

\begin{document}
\maketitle

\begin{abstract}
Wide-baseline matching (WBM) requires integrating geometric understanding, viewpoint changes, fine-grained perception, and occlusion reasoning, making it a challenging testbed for spatial reasoning in multimodal large language models (MLLMs) deployed in physical environments. However, current MLLMs lack systematic evaluation and training frameworks for these capabilities. We introduce \textbf{ReasonMatch-Bench}, a benchmark stratified by viewpoint displacement and matching granularity across indoor, outdoor, and object-centric scenarios, and show that current MLLMs still struggle with fine-grained wide-baseline correspondence: on a difficult 90-sample subset, human annotators achieve 84.0 F1, while the best existing baseline reaches 37.2. To bridge this gap, we build a scalable data-generation pipeline that automatically extracts wide-baseline view pairs from large-scale video-3D corpora, including RGB-D videos and SfM reconstructions, yielding diverse and verifiable supervision. We further propose \textbf{Dynamic Correspondence Reinforcement Learning} (DCRL), which combines Image-Level Viewpoint Progression and Point-Level Correspondence Curriculum to improve WBM training through verifiable rewards without explicit CoT supervision. Extensive experiments show that DCRL substantially improves ReasonMatch-Bench and transfers to related spatial benchmarks, while maintaining general visual understanding performance with modest gains on several benchmarks.
\end{abstract}
 \newline

\section{Introduction}
\label{sec:intro}
Deploying multimodal large language models (M\-L\-L\-Ms)~\cite{wang2024qwen2vl,zhu2025internvl3,liu2024llava15} in the physical world requires more than object recognition or captioning: it requires spatial reasoning across disparate viewpoints. Such reasoning involves geometric understanding~\cite{wang2025moge,wang2025vggt}, viewpoint imagination~\cite{yin2025mindcube,yeh2025seeing}, fine-grained perception like segmentation and detection ~\cite{kirillov2023segment,zhu2025active,zhu2025segagent}, occlusion and topological reasoning~\cite{pang2025manivideo,banerjee2025hot3d,huang2025notvla}, and scale or depth estimation~\cite{hu2024metric3d,bochkovskii2024depth}. Despite rapid progress in MLLMs, how to \emph{train} and \emph{evaluate} these capabilities in a unified, scalable, and verifiable manner remains open.

A central obstacle is data. Curating supervision that truly elicits spatial reasoning is expensive and brittle. Manual annotation rarely captures the full mix of geometry, semantics, and context in a single example, while synthetic setups often struggle to match real-world diversity and verification at scale. This motivates a practical question: can we leverage existing large-scale video-3D data to both \emph{test} and \emph{improve} spatial reasoning in MLLMs with minimal human effort?

We revisit this challenge through the lens of \emph{Wide-Baseline Matching} (WBM)~\cite{schmid1995matching,pritchett1998wide}: deciding whether two views separated by large baselines, strong perspective and appearance changes, repetitive structures, illumination shifts, and semantic occlusions depict the same physical scene element. Classical feature-based pipelines can be effective under small viewpoint changes or dense frame sampling, but they frequently fail in the extreme regime. Humans, on the contrary, still succeed by jointly exploiting geometric regularities, semantic knowledge, and contextual cues. This raises two questions: how well do current MLLMs handle WBM task, and what data and training paradigm can \emph{reliably} improve this ability?

\begin{figure*}[ht]
        \centering
    \includegraphics[width=.95\linewidth]{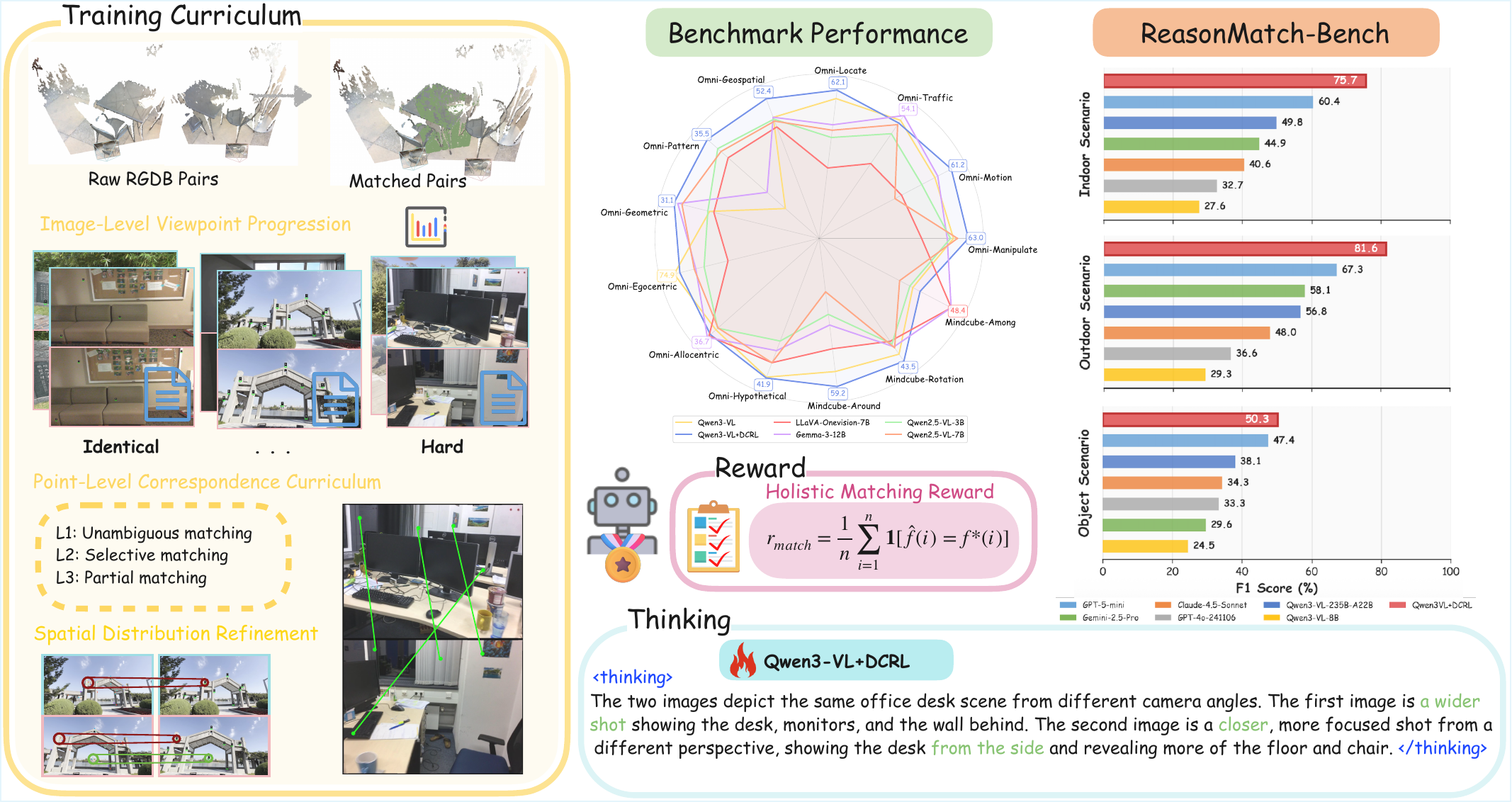}
    \caption{
    \textbf{Wide-Baseline Matching exposes major spatial-reasoning gaps in current MLLMs.}
    Right: even frontier models struggle on ReasonMatch-Bench. Middle: DCRL improves ReasonMatch and transfers to other spatial intelligence benchmarks. Left: Dataset curation pipeline from RGBD datas to different WBM question-answer pairs.
    }
    
    \label{fig:my_pdf}
\end{figure*}

To answer these questions, we introduce \textbf{ReasonMatch-Bench}, a comprehensive benchmark for assessing cross-view spatial reasoning in MLLMs through wide-baseline matching. 
ReasonMatch-Bench stratifies difficulty by viewpoint change magnitude and matching granularity, spanning indoor, outdoor, and object-centric scenarios.
Our study reveals that current MLLMs still struggle on wide-baseline matching. On a difficult 90-sample human-study subset shown in Table~\ref{tab:human_main}, human annotators achieve 84.0 F1, while the best existing baseline reaches 37.2, and smaller models perform worse still. To improve this ability, we introduce a scalable data-generation pipeline that \emph{automatically} harvests wide-baseline view pairs from large-scale video-3D corpora, including RGB-D videos and Structure-from-Motion (SfM) reconstructions, yielding diverse, verifiable supervision across scenarios and matching granularities.

Leveraging the verifiable nature of wide-baseline matching, we optimize MLLMs via \emph{reinforcement learning with verifiable rewards} (RLVR)~\cite{guo2025deepseek,wen2025reinforcement,zhong2025omni}: the model is rewarded according to its matching accuracy, enabling it to improve spatial reasoning without explicit reasoning supervision. To train this capability effectively, we introduce \emph{Dynamic Correspondence Reinforcement Learning} (DCRL), which combines Image-Level Viewpoint Progression and Point-Level Correspondence Curriculum to perform a sample-efficient training process. Our experiments show that DCRL improves ReasonMatch-Bench substantially to 70.5\% F1 score, outperforming both open-source and closed-source baselines including GPT-5-mini (57.9\%) and Gemini-2.5-Pro (42.8\%), and also transfers to related spatial benchmarks including OmniSpatial (+5.27\%) and MindCube (+3.51\%) while maintaining general visual understanding performance with modest gains on several benchmarks.

Our work makes the following contributions:
\begin{itemize}
    \item We introduce \textbf{ReasonMatch-Bench}, a comprehensive benchmark for evaluating spatial reasoning in MLLMs through wide-baseline matching, spanning indoor, outdoor, and object-centric scenarios with stratified difficulty levels.
    
    \item We propose \textbf{Dynamic Correspondence Reinforcement Learning (DCRL)}, a curriculum-based framework with dual-level adaptive curricula that enables MLLMs to progressively master complex spatial reasoning through verifiable rewards.
    
    \item We show that DCRL improves ReasonMatch-Bench, transfers to related spatial benchmarks, and does not degrade general visual understanding.
\end{itemize}
 \section{Related Work}
\label{sec:related_work}

{\bf Spatial Reasoning in MLLMs}.
Evaluating and eliciting complex spatial reasoning in MLLMs remains an open challenge~\cite{dongfang2025multimodal,jia2025omnispatial,wang2025site,ma20253dsrbench,yang2025thinking}.
Existing benchmarks such as OmniSpatial~\cite{jia2025omnispatial} and VSI-Bench~\cite{yang2025thinking} assess various facets of spatial understanding, yet individual samples typically probe isolated capabilities such as relative positioning or viewpoint prediction, rather than requiring integrated reasoning across geometry, semantics, and context. 
On the training side, methods like SAT~\cite{ray2024sat}, RoboSpatial~\cite{song2025robospatial}, and RoboRefer~\cite{zhou2025roborefer} focus primarily on visual grounding or simple relational reasoning, staying mainly on textual reasoning and MCQ evaluation.
Multi-SpatialMLLM~\cite{xu2025multi} explores correspondence matching but is limited to small viewpoint changes, restricted task formats (e.g., multiple-choice), and supervised fine-tuning (SFT) alone, which may not sufficiently elicit deeper spatial reasoning.
In contrast, our work starts from \emph{wide-baseline matching} (WBM), a fundamental yet challenging visual task that naturally demands complex spatial reasoning.
Drawing inspiration from the success of reinforcement learning in DeepSeek-R1~\cite{guo2025deepseek} and leveraging the inherent verifiability of matching tasks through geometric constraints, we employ reinforcement learning to enable MLLMs to autonomously explore and acquire complex spatial reasoning capabilities.
This approach allows the model to discover reasoning strategies beyond what supervised annotations can provide, with the goal of improving performance on broader spatial intelligence tasks.

\noindent 
{\bf Wide Baseline View Matching}. 
Wide baseline view matching refers to the task of finding correspondences between two or more views of a scene captured from significantly different viewpoints. This problem is fundamental to many computer vision applications, including 3D reconstruction and re-localization \cite{jin2020image}.
Traditional approaches relied on a pipeline of extracting handcrafted local features (e.g., SIFT \cite{lowe2004distinctive}, SURF \cite{bay2006surf}, ORB \cite{rublee2011orb}), matching them, and using a robust estimator like RANSAC \cite{fischler1981random} to find the epipolar geometry \cite{hartley1994projective}. 
Subsequent research improved each component, introducing learned descriptors \cite{mishchuk2017working, tian2019sosnet}, end-to-end feature networks \cite{detone2018superpoint, dusmanu2019d2, revaud2019r2d2}, and advanced robust estimators \cite{barath2018graph, barath2019magsac}.
However, these feature-centric methods frequently fail in the \textit{extreme regime} \cite{jin2020image}. The severe changes in perspective, illumination, and occlusion inherent to this task demand robust reasoning about geometry, semantics, and context, which these methods lack.
 \section{Method}
\label{sec:method}

\begin{figure*}[t]
    \centering
    \includegraphics[width=\linewidth]{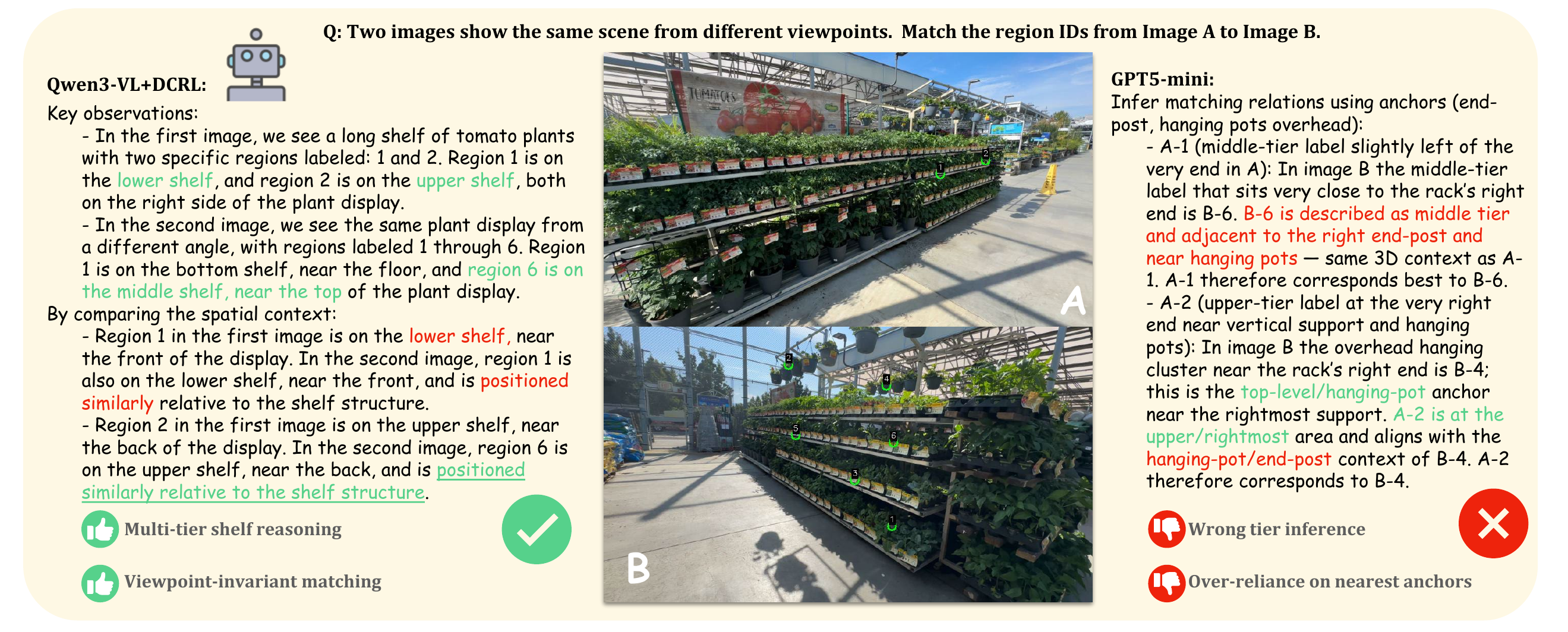}
    \caption{
    Comparison of cross-view region matching. 
    Our method correctly preserves global spatial consistency across viewpoints, 
    reasoning over multi-tier shelf structure and stable anchors. 
    GPT-5-mini fails to perform viewpoint-consistent alignment and confuses shelf tiers, 
    leading to incorrect cross-view correspondences.
    \textbf{Note:} 
    \textcolor{green!60!black}{Green highlights indicate correct spatial reasoning;} 
    \textcolor{red!70!black}{red highlights indicate incorrect reasoning.}
    }

    \label{fig:reasoning_case_study}
\end{figure*}

We now describe our approach for eliciting complex spatial reasoning in MLLMs through Wide-Baseline Matching.
Our method comprises three main components: task formulation for MLLMs to perform WBM,
a scalable dataset and benchmark generation pipeline, a reinforcement learning framework using verifiable rewards combined with a curriculum strategy to progressively enhance spatial reasoning capabilities.

\subsection{Task Formulation for Wide-Baseline Matching}

\textbf{Problem Definition.}
We define the cross-view matching problem as follows. 
Given two images $I_1, I_2$ depicting the same 3D scene but captured from different viewpoints, 
let each camera be parameterized by its intrinsic matrix $\mathbf{K}_i$ and extrinsic parameters $(\mathbf{R}_i, \mathbf{t}_i)$, 
where $\mathbf{R}_i \in SO(3)$ and $\mathbf{t}_i \in \mathbb{R}^3$ denote the rotation and translation with respect to a common world coordinate frame. 
For a 3D point $\mathbf{X} \in \mathbb{R}^3$, the image projection is given by the standard camera model in homogeneous coordinates:
$\pi_i(\mathbf{X}) = \mathbf{K}_i [\mathbf{R}_i \,|\, \mathbf{t}_i] \mathbf{X}$.
The goal is to predict a set of correspondences
\begin{equation}
\mathcal{M} = \{(\mathbf{x}_i, \mathbf{x}'_i)\}_{i=1}^{N},
\label{eq:correspondence-set}
\end{equation}
such that each pair corresponds to the same 3D point in the scene, i.e.,
$\mathbf{x}_i = \pi_1(\mathbf{X}_i)$ and $\mathbf{x}'_i = \pi_2(\mathbf{X}_i)$.
Given sufficient and accurate correspondences, the relative camera pose $(\mathbf{R}, \mathbf{t})$ between the two views can be recovered via the epipolar constraint.  
Classical matching methods directly predict a large set of correspondences $\mathcal{M}$ by computing appearance similarity between local features~\cite{lowe2004distinctive,bay2006surf,rublee2011orb}, followed by geometric verification~\cite{hartley1994projective,schonberger2016structure}.

\noindent \textbf{Text-driven correspondence reasoning.}
Unlike classical matchers that output a continuous score matrix 
$S \in \mathbb{R}^{n \times m}$,
the MLLM performs matching in a discrete, language-mediated manner.
Given two pre-marked point sets
$\mathcal{X}=\{\mathbf{x}_i\}_{i=1}^{n}$ and 
$\mathcal{Y}=\{\mathbf{y}_j\}_{j=1}^{m}$,
the model receives $(I_1,\mathcal{X};\, I_2,\mathcal{Y})$ as input, 
with visual prompts indicating point indices, and produces a textual mapping
\begin{equation}
\hat{f}: \{1,\dots,n\} \rightarrow \{1,\dots,m\} \cup \{\varnothing\},
\label{eq:mllm-mapping}
\end{equation}
where $\hat{f}(i)=j$ means that $\mathbf{x}_i$ in $I_1$ corresponds to $\mathbf{y}_j$ in $I_2$,
and $\hat{f}(i)=\varnothing$ denotes no confident match.
The predicted correspondence set is
\begin{equation}
\widehat{\mathcal{M}}=\{(\mathbf{x}_i,\mathbf{y}_{\hat{f}(i)}) \mid 
\hat{f}(i)\neq\varnothing\}.
\label{eq:predicted-correspondences}
\end{equation}

Conceptually, this process can be viewed as a \emph{partial bipartite matching}~\cite{burkard2012assignment,gold2002graduated} between 
the two point sets $\mathcal{X}$ and $\mathcal{Y}$, 
where each point may correspond to at most one counterpart or remain unmatched 
due to occlusion or limited overlap. 
This formulation treats the MLLM as a reasoning engine that performs symbolic association between visual entities rather than continuous feature matching,
allowing it to integrate geometric, semantic, and contextual cues through complex spatial reasoning.

\subsection{Dataset Generation Pipeline}

We now describe how to construct data samples $(I_1, \mathcal{X}; I_2, \mathcal{Y})$ with ground-truth correspondences.
Each point set consists of matchable and distractor subsets:
\begin{equation}
\mathcal{X} = \mathcal{X}^{\text{match}} \cup \mathcal{X}^{\text{dist}}, \quad
\mathcal{Y} = \mathcal{Y}^{\text{match}} \cup \mathcal{Y}^{\text{dist}},
\label{eq:point-partition}
\end{equation}
where $\mathcal{X}^{\text{match}}, \mathcal{Y}^{\text{match}}$ have valid correspondences and $\mathcal{X}^{\text{dist}}, \mathcal{Y}^{\text{dist}}$ are distractors.
Our data generation pipeline focuses on obtaining image pairs $(I_1, I_2)$ with verified correspondences that serve as the foundation for constructing these point sets.

\noindent \textbf{Image pair selection and correspondence extraction.}
We source image pairs from diverse RGB-D datasets (CO3D~\cite{reizenstein2021common}, uCO3D~\cite{liu24uco3d}, ScanNet~\cite{dai2017scannet}) and RGB videos with SfM reconstructions (RealEstate10k~\cite{zhou2018stereo}, DL3DV~\cite{ling2024dl3dv}).
For RGB-D data, we obtain correspondences via geometric reprojection: each pixel in $I_1$ with valid depth is back-projected to 3D and reprojected into $I_2$, then verified using depth consistency and photometric consistency checks (see supplementary for details).
For SfM data, we extract correspondences from shared 3D landmarks in COLMAP reconstructions~\cite{schonberger2016structure}, which have already passed geometric verification.
This process yields a dense correspondence set $\mathcal{M}$ with thousands of matches per pair.

\noindent \textbf{Viewpoint difficulty quantification.}
We quantify the viewpoint change between $(I_1, I_2)$ using an overlap score $\omega \in [0,1]$: for RGB-D pairs, $\omega$ measures the fraction of successfully matched pixels; for SfM pairs, $\omega$ reflects the proportion of shared 3D landmarks (details in supplementary). We use this score for source-aware difficulty stratification rather than direct cross-source comparison.
We define the viewpoint-change magnitude as $\Delta_v = 1 - \omega$, which increases with baseline distance and occlusion.
This metric enables us to stratify pairs by viewpoint difficulty and supports curriculum-based data organization.

\noindent \textbf{Constructing the verified correspondence pool.}
The raw dense matches $\mathcal{M}$ are unsuitable for direct use: they cause severe visual overlap when marked on images and exceed practical input limits for MLLMs.
We therefore apply clustering-based spatial filtering to subsample $\mathcal{M}$ into a moderate-sized verified pool $\mathcal{P} = \{(\mathbf{p}_i^1, \mathbf{p}_i^2)\}_{i=1}^{N_p}$ with typically $N_p \in [10, 50]$ spatially well-separated correspondences per pair.
Specifically, we cluster matches in joint image-coordinate space and retain one representative per cluster to ensure adequate spacing for visual prompting.
The final preprocessed samples $(I_1, I_2, \mathcal{P})$ provide a high-quality correspondence pool from which matchable and distractor points can be flexibly sampled for various training and evaluation scenarios.

\subsection{DCRL: Dynamic Correspondence Reinforcement Learning}
\label{sec:curriculum_def}

Since WBM admits verifiable rewards but remains difficult under extreme viewpoint changes, we optimize with RLVR on our preprocessed samples $(I_1, I_2, \mathcal{P})$ to enable MLLMs to autonomously develop spatial reasoning capabilities through \emph{exploration} rather than merely imitating supervised demonstrations.

However, directly training on extreme matching scenarios can lead to inefficient exploration and poor convergence. We therefore introduce a progressive curriculum that decomposes the difficulty along two complementary dimensions: \emph{image-level} viewpoint progression that gradually increases geometric transformation complexity, and \emph{point-level} correspondence progression that adaptively adjusts the number and spatial distribution of matchable points and distractors. This hierarchical decomposition enables the model to build spatial reasoning capabilities incrementally, mastering simpler configurations before tackling extreme scenarios.

DCRL comprises: (1)
a holistic matching reward that evaluates all query regions including unmatched points, encouraging comprehensive spatial reasoning; (2) an image-level progression that stages training by viewpoint divergence; (3) a point-level curriculum with two sub-dimensions—correspondence cardinality and spatial distribution—that dynamically adjusts task construction within each viewpoint stage.

\paragraph{Holistic Matching Reward.}
Traditional partial bipartite matching evaluates only matched pairs, ignoring unmatched points. To encourage comprehensive spatial reasoning over \emph{all} regions including occluded or out-of-view areas, we explicitly assign a dummy target ($\varnothing$ in Eq.~\eqref{eq:mllm-mapping}) to unmatched points and reward correct ``no match'' predictions. This design eliminates objective ambiguity and prevents the model from focusing solely on easily matchable salient features, instead requiring deliberate reasoning about viewpoint-dependent visibility and geometric constraints across the entire scene.

Given predicted mapping $\hat{f}$ and ground-truth $f^*$ over $n$ query regions, we define matching correctness as:
\begin{equation}
r_{\text{match}} = \frac{1}{n}\sum_{i=1}^{n} \mathbbm{1}\left[\hat{f}(i) = f^*(i)\right],
\end{equation}
which measures prediction accuracy over all regions, rewarding correct predictions including unmatched regions ($\varnothing$). We additionally incorporate a format compliance component to ensure well-formed outputs, yielding final reward $r = w_f \cdot r_{\text{format}} + w_m \cdot r_{\text{match}}$. 
Importantly, $r_{\text{match}}$ serves not only as the training signal for policy optimization but also as the control signal for dynamically adapting task difficulty across the curriculum dimensions described below.
\paragraph{Image-Level Viewpoint Progression.}
Rather than training on randomly shuffled image pairs, we partition the dataset by viewpoint overlap score $\omega$ to enable gradual adaptation to geometric complexity. Specifically, the dataset is organized into bins $\{\mathcal{D}_s\}_{s=1}^S$ by overlap intervals $[\underline{\omega}_s, \overline{\omega}_s]$, where bin $s=1$ contains high-overlap pairs with minimal viewpoint change and bin $s=S$ contains extreme viewpoint divergence. Training proceeds sequentially through these bins: once sustained performance on $\mathcal{D}_s$ exceeds a threshold, we advance to $\mathcal{D}_{s+1}$ and permanently exclude the easier bin from the training set.
This staged progression offers two key benefits. In early training, the model focuses on simpler geometric transformations, rapidly building foundational spatial reasoning capabilities and achieving faster convergence. In later stages, the model leverages its established understanding to tackle extreme viewpoint changes, where the more challenging scenarios provide richer learning signals and greater information gain. By filtering out mastered configurations, we maintain training efficiency while progressively pushing the frontier of the model's spatial reasoning capabilities.

\paragraph{Point-Level Correspondence Curriculum.}
A key feature of our approach is that point sets $\mathcal{X}, \mathcal{Y}$ are not pre-marked offline on images. Instead, we dynamically sample them from the verified pool via $\mathcal{X}, \mathcal{Y} = g(\mathcal{P})$, where the sampling strategy $g$ adapts to control task difficulty. Within each viewpoint stage, we modulate task complexity along two sub-dimensions: (1)~\emph{Cardinality adaptation} adjusts the number of matchable points and distractors to vary selection ambiguity; (2)~\emph{Spatial distribution refinement} modulates the spatial arrangement of sampled points to influence local versus global reasoning demands.

\noindent\textbf{Cardinality adaptation.} 
When sampling point sets $\mathcal{X}, \mathcal{Y}$ from the verified pool $\mathcal{P}$, we dynamically adjust the number of matchable and distractor points to control task difficulty.
Recall the point partition in Eq.~\ref{eq:point-partition}. We construct three progressive difficulty levels:

\begin{itemize}
\item \textbf{Unambiguous matching (L1):} $\mathcal{X}^{\text{dist}} = \mathcal{Y}^{\text{dist}} = \varnothing$ with $|\mathcal{X}^{\text{match}}| = |\mathcal{Y}^{\text{match}}| = n$. One-to-one correspondence eliminates selection ambiguity, allowing the model to focus on geometric transformation understanding. The model learns to reason about appearance changes, occlusion boundaries, and 3D structure without distractor interference.

\item \textbf{Selective matching (L2):} $\mathcal{X}^{\text{dist}} = \varnothing$, $|\mathcal{Y}^{\text{dist}}| > 0$. Introduces the challenge of selecting correct matches from multiple candidates in $\mathcal{Y}$. This simulates asymmetric scene coverage where one view observes additional regions, requiring the model to distinguish geometrically consistent correspondences from visually similar but incorrect candidates.

\item \textbf{Partial matching (L3):} Both $|\mathcal{X}^{\text{dist}}| > 0$ and $|\mathcal{Y}^{\text{dist}}| > 0$. Models realistic scenarios with bidirectional occlusion and incomplete overlap. The model must explicitly reason about which regions are visible in both views versus occluded or out-of-frame, integrating geometric constraints with semantic understanding of scene structure and viewpoint-dependent visibility.
\end{itemize}
The curriculum dynamically promotes to higher levels as performance improves and demotes upon performance degradation, adapting task complexity to the model's evolving capabilities.

\noindent\textbf{Spatial distribution refinement.} 
Beyond cardinality, the spatial distribution of matchable points sampled from $\mathcal{P}$ significantly impacts task difficulty. Points that are too densely clustered become difficult to distinguish or form fixed patterns that enable matching without visual context reasoning. Conversely, overly sparse points limit the model to coarse object-level understanding without learning fine-grained spatial relationships. We therefore dynamically control spatial distribution through clustering radius and sampling strategies to progressively refine spatial reasoning capabilities.

\begin{table*}[ht]
    \centering
    \caption{
    Performance comparison on \textbf{ReasonMatch-Bench}. We report F1, Precision, and Recall scores across three scenarios (Indoor, Outdoor, Object) and three difficulty levels.}
    \label{tab:model_comparison_percent}
    \footnotesize 
    \setlength{\tabcolsep}{4pt} 
    \sisetup{detect-weight, mode=text}
    
    \begin{tabular}{l *{12}{S[table-format=2.1]}}
    \toprule
    \textbf{Model} & 
    \textbf{F1} & 
    \textbf{Precision} & 
    \textbf{Recall} & 
    \multicolumn{3}{c}{\textbf{Indoor}} & 
    \multicolumn{3}{c}{\textbf{Outdoor}} & 
    \multicolumn{3}{c}{\textbf{Object}} \\
    \cmidrule(r){5-7} \cmidrule(lr){8-10} \cmidrule(l){11-13}
    & & & & 
    \textbf{L1} & \textbf{L2} & \textbf{L3} & 
    \textbf{L1} & \textbf{L2} & \textbf{L3} & 
    \textbf{L1} & \textbf{L2} & \textbf{L3} \\
    \midrule
    GPT-5-mini              & 57.9 & 56.9 & 59.4 & 68.2 & 65.3 & 47.0 & 75.8 & 75.0 & 51.4 & 45.3 & 60.4 & 27.8 \\
    GPT-5-Chat              & 51.5 & 50.6 & 52.8 & 52.4 & 62.5 & 39.3 & 61.8 & 75.1 & 44.9 & 44.1 & 57.2 & 28.6 \\
    GPT-4o-241106           & 33.5 & 32.7 & 34.7 & 31.9 & 43.7 & 21.7 & 33.9 & 54.0 & 23.3 & 31.9 & 42.2 & 19.9 \\
    Gemini-2.5-Pro          & 42.8 & 42.4 & 43.4 & 48.3 & 49.3 & 36.7 & 63.4 & 64.5 & 46.6 & 29.5 & 35.9 & 19.0 \\
    Claude-4.5-Sonnet       & 41.7 & 43.7 & 41.1 & 40.9 & 50.2 & 29.9 & 50.3 & 56.9 & 37.4 & 33.9 & 43.4 & 19.4 \\
    Claude-4.1-Opus         & 33.7 & 37.2 & 32.2 & 31.0 & 36.0 & 23.8 & 45.1 & 45.3 & 31.6 & 26.3 & 37.2 & 16.1 \\
    Claude-4-Sonnet         & 34.8 & 34.2 & 35.5 & 33.0 & 43.7 & 21.3 & 45.0 & 54.5 & 29.8 & 30.8 & 44.1 & 18.3 \\
    \midrule
    Qwen3-VL-235B~\cite{bai2025qwen3}        & 49.2 & 50.7 & 48.7 & 52.0 & 60.5 & 35.8 & 58.8 & 71.1 & 41.7 & 35.1 & 49.4 & 22.1 \\
    Qwen3-VL-8B-Instruct~\cite{bai2025qwen3} & 27.5 & 27.1 & 29.1 & 25.5 & 38.8 & 17.6 & 30.4 & 39.5 & 18.8 & 21.1 & 34.5 & 11.5 \\
    \midrule
    \bfseries Qwen3-VL-8B + DCRL     & \bfseries 70.5 & \bfseries 70.3 & \bfseries 71.1 & \bfseries 84.6 & \bfseries 75.1 & \bfseries 67.0 & \bfseries 90.9 & \bfseries 80.2 & \bfseries 73.6 & \bfseries 45.6 & \bfseries 63.1 & \bfseries 33.7 \\
    \rowcolor{gray!15}
    $\Delta$ vs. Qwen3-VL-8B-Instruct & +43.0 & +43.2 & +42.0 & +59.0 & +36.3 & +49.4 & +60.5 & +40.7 & +54.8 & +24.5 & +28.6 & +22.2 \\
    \bottomrule
    \end{tabular}
\end{table*}

Specifically, we employ a cluster-based sampling approach where correspondences are first grouped by spatial proximity, then representatives are selected per cluster. We progress through three stages: (1)~\emph{Maximally sparse sampling}: selecting one point per cluster with large clustering radius, producing globally distributed points that require object-level reasoning; (2)~\emph{Moderate clustering}: reducing the clustering radius to allow multiple points per region, introducing finer spatial structure; (3)~\emph{Dense sampling}: transitioning to random sampling with minimal spacing constraints, requiring the model to reason about detailed geometric relationships and subtle appearance variations at fine granularity. This progression gradually eliminates spatial cues that facilitate matching, forcing the model to develop comprehensive geometric understanding rather than relying on global landmark distribution alone.

This two-level hierarchy—image-level viewpoint filtering as the outer loop and point-level adaptive construction as the inner loop—enables efficient exploration by aligning task difficulty with the model's evolving spatial reasoning capabilities.

\section{Experiments}
\label{sec:experiments}

\subsection{Dataset and Benchmark Statistics}

\noindent 
{\bf TestSet Composition and Balance.}
To evaluate WBM in MLLMs, we curate 2,810 image pairs from our 220k-pair corpus as \textbf{ReasonMatch-Bench}. The benchmark balances data sources (ScanNet 27.7\%, uCO3D 28.0\%, DL3DV 27.0\%, RE10K 17.2\%), task levels (L1 32.5\%, L2 36.8\%, L3 30.7\% as defined in Sec.~\ref{sec:curriculum_def}), and scene types (indoor 55.1\%, object 28.0\%, outdoor 16.9\%). This balance extends to cross-dimensional distributions: within each dataset, task composition remains approximately uniform. It ensures that evaluation metrics reflect broad spatial reasoning performance rather than biases toward particular sources or configurations. See the supplementary material for detailed statistics.

\subsection{Implementation Details}   

\noindent 
{\bf RLVR Configuration and Training Setup. }
We apply GRPO on Qwen3-VL-8B-Instruct~\cite{bai2025qwen3} with a group size $G=32$ to ensure sufficient variance reduction and exploration of diverse reasoning paths during policy optimization. The effective batch size is 16$\times$32 trajectories per update. The KL loss coefficient is set to $\beta=0.005$. Each generated prediction is capped at 5120 tokens and the temperature is set $T=1.0$ for the policy rollouts to maintain sufficient exploration without introducing excessive randomness. We use AdamW optimizer~\cite{loshchilov2017decoupled} with a linear warmup over the first 10 steps, and a constant learning rate of $ 10^{-6}$. 

\noindent 
{\bf Reward and Curriculum.}
The reward function uses weights $(w_f, w_m) = (1.0, 1.0)$ as described in Section~\ref{sec:method}, with format compliance and matching correctness weighted equally. 
For curriculum configuration, we organize training along three hierarchical dimensions. At the viewpoint level, we partition the dataset into 10 overlap bins and advance to the next bin once the average accuracy reward exceeds 0.8 over a sliding window of 20 training steps. The cardinality adaptation strategy and spatial correspondence distribution settings are shown in the supplement.

\begin{table*}[ht]
\centering
\small
\renewcommand{\arraystretch}{1.15}
\setlength{\tabcolsep}{2pt}
\caption{
Results on the \textbf{OmniSpatial} benchmark. Accuracy (\%) is reported for overall and the four major reasoning dimensions. 
}
\label{tab:omnispatial_summary}
\resizebox{.859\textwidth}{!}{
\begin{tabular}{lcccccc}
\toprule
Model & Size & Overall & Dynamic Reasoning & Spatial Interaction & Complex Logic & Perspective Taking \\
\midrule
GPT-4-turbo~\cite{openai2023gpt4} & -- & 34.06 & 38.39 & 36.49 & 24.80 & 33.69 \\
Gemini-2.5-flash-preview-05-20~\cite{gemini2023anil} & -- & 52.12 & 63.59 & 67.46 & 35.67 & 43.10 \\
LLaVA-1.5-vicuna-7B~\cite{liu2024llava15} & 7B & 34.97 & 35.38 & 35.13 & 25.99 & 38.82 \\
InternVL3-78B~\cite{zhu2025internvl3} & 78B & 49.33 & 63.24 & 55.61 & 29.23 & 44.93 \\
Qwen2.5-VL~\cite{bai2025qwen25vltechnicalreport} & 7B & 39.25 & 46.30 & 30.06 & 35.65 & 39.68 \\
\midrule
Qwen3-VL~\cite{bai2025qwen3} & 8B & 43.60 & 51.90 & 51.90 & 24.40 & 42.50 \\
\textbf{DCRL} & 8B & \textbf{48.87} & \textbf{61.48} & \textbf{55.33} & \textbf{32.78} & \textbf{43.21} \\
\bottomrule
\end{tabular}
}
\end{table*}

\subsection{Main Results on ReasonMatch-Bench}
\paragraph{Performance Analysis across Scene Types and Task Levels.}
Table~\ref{tab:model_comparison_percent} presents a comprehensive comparison across three scene types and three task difficulty levels. Additional qualitative analysis of model CoT outputs is provided in the supplement. Several notable patterns emerge from this evaluation.

\textbf{Our method achieves the best overall result and strong gains in challenging settings.} The improvement is particularly pronounced in difficult scenarios: while maintaining strong performance on easier outdoor scenes, our model also shows robust performance on complex indoor and instance-level matching tasks where baseline models struggle significantly.

\textbf{Scene-related difficulty reveals dataset characteristics.} Outdoor scenes prove most tractable across all models, with even baseline systems achieving reasonable performance on L1 tasks. Indoor scenes present moderate complexity, where our method maintains substantial advantages over strong proprietary baselines on many difficult configurations. Instance-level matching emerges as the most challenging scenario—baseline models show dramatic performance degradation, particularly on L3 tasks, while our approach maintains relatively stable performance. This pattern reflects the fundamental challenge of object-centric matching: isolated objects lack environmental context that aids correspondence reasoning in full scenes.

\begin{table}[b]
\centering
\small
\setlength{\tabcolsep}{4pt}
\caption{Human study on a 90-sample high-divergence subset. We report F1 only; the full precision/recall/F1 breakdown is provided in the supplement.}
\label{tab:human_main}
\begin{tabular}{lcccc}
\toprule
\textbf{Method} & \textbf{Overall} & \textbf{DL3DV} & \textbf{RE10K} & \textbf{uCO3D} \\
\midrule
GPT-5-mini & 37.2 & 35.9 & 49.7 & 25.8 \\
Gemini-2.5-Pro & 29.5 & 26.5 & 44.1 & 18.0 \\
Claude-4.5-Sonnet & 24.0 & 22.2 & 30.5 & 19.2 \\
Qwen3-VL-235B~\cite{bai2025qwen3} & 29.9 & 25.3 & 45.7 & 18.7 \\
\textbf{DCRL} & \textbf{52.0} & \textbf{57.7} & \textbf{70.6} & \textbf{27.8} \\
\midrule
\rowcolor{gray!15}
\textbf{Human} & \textbf{84.0} & \textbf{93.5} & \textbf{94.7} & \textbf{62.1} \\
\bottomrule
\end{tabular}
\end{table}

To calibrate benchmark difficulty against human performance, we additionally evaluate annotators on the 90 largest-view-divergence samples from DL3DV, RE10K, and uCO3D including indoor, outdoor and object level scenes. Table~\ref{tab:human_main} shows that humans achieve 84.0 F1 overall, compared with 52.0 for DCRL. The remaining gap is especially large on object-centric uCO3D (62.1 vs. 27.8), indicating that challenging wide-baseline correspondence remains far from solved even after DCRL training.

\begin{table}[ht]
\centering
\small
\renewcommand{\arraystretch}{1}
\caption{
Results on \textbf{MindCube} and \textbf{SAT Real} benchmarks. 
Accuracy (\%) for spatial reasoning tasks.
}
\label{tab:spatial_benchmarks}
\resizebox{0.95\linewidth}{!}{
\begin{tabular}{lccccc}
\toprule
\multirow{2}{*}{\textbf{Model}} & \multicolumn{4}{c}{\textbf{MindCube}} & \textbf{SAT} \\
\cmidrule(lr){2-5} \cmidrule(lr){6-6}
 & Overall & Rotation & Among & Around & Real \\
\midrule
\multicolumn{6}{l}{\textit{Open-Weight Multi Image Models}} \\
LongVA-7B~\cite{zhang2024longva} & 29.46 & 35.89 & 29.55 & 24.88 & -- \\
InternVL2.5-8B~\cite{chen2024expanding} & 18.68 & 36.45 & 18.20 & 13.11 & -- \\
Qwen2.5-VL-7B~\cite{bai2025qwen25vltechnicalreport} & 29.26 & 38.76 & 29.50 & 21.35 & 56.33 \\
Idefics3-8B-Llama3~\cite{laurencon2024idefics} & 35.86 & 35.15 & 35.94 & 35.49 & -- \\
\midrule
\multicolumn{6}{l}{\textit{Proprietary Model}} \\
GPT-4o~\cite{openai2024gpt4o} & 38.81 & 32.65 & 40.17 & 29.16 & 57.50 \\
\midrule
\multicolumn{6}{l}{\textit{Spatial Models}} \\
RoboBrain~\cite{ji2025robobrain} & 37.38 & 35.80 & 38.28 & 29.53 & -- \\
SpatialVLM~\cite{chen2024spacemantis} & 22.81 & 37.65 & 21.26 & 29.39 & -- \\
\midrule
\multicolumn{6}{l}{\textit{Our Models}} \\
Qwen3-VL-8B~\cite{bai2025qwen3} & 40.01 & 53.20 & 41.00 & 34.33 & 70.00 \\
\textbf{DCRL} & \textbf{43.52} & \textbf{59.20} & \textbf{43.50} & \textbf{37.00} & \textbf{75.30} \\
\bottomrule
\end{tabular}
}
\end{table}

\textbf{Qualitative analysis of failure modes.} Examining model outputs reveals distinct reasoning patterns across baselines. 
Gemini-2.5-Pro demonstrates accurate point-level descriptions, providing detailed local appearance characterizations for each annotated region. However, these descriptions, while locally correct, lack global specificity and discriminative power within the full scene context—descriptions like ``white wall region'' or ``wooden surface'' may accurately characterize local appearance but fail to uniquely identify the target point when multiple similar regions exist. This inability to leverage holistic scene understanding and 3D spatial relationships leads the model to perform ambiguous local feature matching rather than geometric correspondence reasoning.
The Qwen3-VL series exhibits complementary strengths and weaknesses: these models show strong awareness of viewpoint changes and can reason about cross-view geometric transformations effectively. However, they suffer from frequent visual label misidentification and reasoning-answer inconsistencies, where the model's Chain-of-Thought reasoning arrives at correct correspondences but the final formatted output contradicts this reasoning. We attribute this to Qwen3-VL's prior training on spatial intelligence tasks providing geometric intuition, but insufficient exposure to fine-grained cross-view scenarios and multi-image contexts leads to persistent hallucination issues when parsing dense visual annotations.

\begin{table}[t]
\centering

\caption{Performance on general visual understanding benchmarks, measured using lmms-eval~\cite{lmms_eval}.}
\label{tab:general_results}
\resizebox{\linewidth}{!}{
\begin{tabular}{lccccccc}
\toprule
Model & MME-RealWorld & MMStar & RealWorldQA & V*Bench \\
\midrule
Qwen3-VL-8B~\cite{bai2025qwen3}  & 62.8 & 59.8 & 69.5 & 84.8 \\
\textbf{DCRL} & \textbf{63.8} & \textbf{62.5} & \textbf{70.5} & \textbf{85.9} \\
\bottomrule
\end{tabular}
}
\end{table}

\subsection{Generalization on other Spatial and Visual Understanding Benchmarks}

To evaluate transfer beyond ReasonMatch-Bench and check whether spatial training affects broader vision-language performance, we test our model on both spatial intelligence benchmarks and general vision-language benchmarks. We compare our model against the base model on three spatial intelligence benchmarks: OmniSpatial~\cite{jia2025omnispatial}, SAT~\cite{ray2024sat}, and MindCube~\cite{yin2025mindcube}. We also report results on four general visual understanding benchmarks: MMStar~\cite{chen2024we}, RealWorldQA~\cite{realworldqa2024}, MME-RealWorld~\cite{zhang2024mme}, and V*Bench~\cite{wu2024v}.

\begin{table}[ht]
\centering
\small
\caption{Performance of RL and SFT training objectives.}
\label{tab:key_ablations}
\resizebox{0.9\linewidth}{!}{
\begin{tabular}{lcccc}
\toprule
Method & OmniSpatial & MindCube & SAT & ReasonMatch \\
\midrule
Base & 43.6 & 40.0 & 70.0 & 27.5 \\
SFT & 42.6 & 45.1 & 41.3 & 51.0 \\
\textbf{DCRL} & \textbf{48.9} & \textbf{43.5} & \textbf{75.3} & \textbf{70.5} \\
\bottomrule
\end{tabular}
}
\end{table}

\paragraph{Performance on Spatial Intelligence and General Visual Benchmarks.}

Table~\ref{tab:spatial_benchmarks} and Table~\ref{tab:omnispatial_summary} present results on three spatial intelligence benchmarks. Our model improves over the base model on the reported spatial benchmarks. These gains suggest positive transfer from cross-view matching to related spatial benchmarks beyond the specific matching task. Examining OmniSpatial's sub-categories reveals heterogeneous transfer patterns. Dynamic Reasoning ($9.6\%$) and Complex Logic ($8.38\%$) show the largest gains among the reported OmniSpatial sub-categories, while Spatial Interaction exhibits moderate gains ($3.4\%$) and Perspective Taking remains nearly stable. One possible explanation is the composition of our training data: many indoor scenes come from room navigation videos, which often include camera rotation and egocentric motion and may therefore better match Complex Logic tasks (3D mental rotation, geometric pattern completion) and Dynamic Reasoning tasks (motion prediction, viewpoint changes). This interpretation is consistent with the MindCube results, where the Rotation sub-task evaluating spatial reasoning under viewpoint rotations exhibits the largest gain ($6.0\%$), notably exceeding improvements on Among and Around sub-tasks.

Table~\ref{tab:general_results} shows no degradation on the reported general visual understanding benchmarks, together with modest gains on all four, suggesting that this spatial training does not harm general visual performance in our evaluation.

\subsection{Analysis and Ablation Studies}

Trained on a CoT-annotated WBM dataset, SFT substantially improves in-domain ReasonMatch performance over the base model, but transfers inconsistently on other benchmarks. In contrast, DCRL improves all reported spatial benchmarks and outperforms SFT by +19.5 on ReasonMatch and +34.0 on SAT. This contrast in Table~\ref{tab:key_ablations} suggests that teacher-forced imitation can overfit to correspondence patterns, whereas reinforcement learning with verifiable rewards develops more transferable spatial reasoning.

\begin{figure}[h]
\centering
\includegraphics[width=0.9\linewidth]{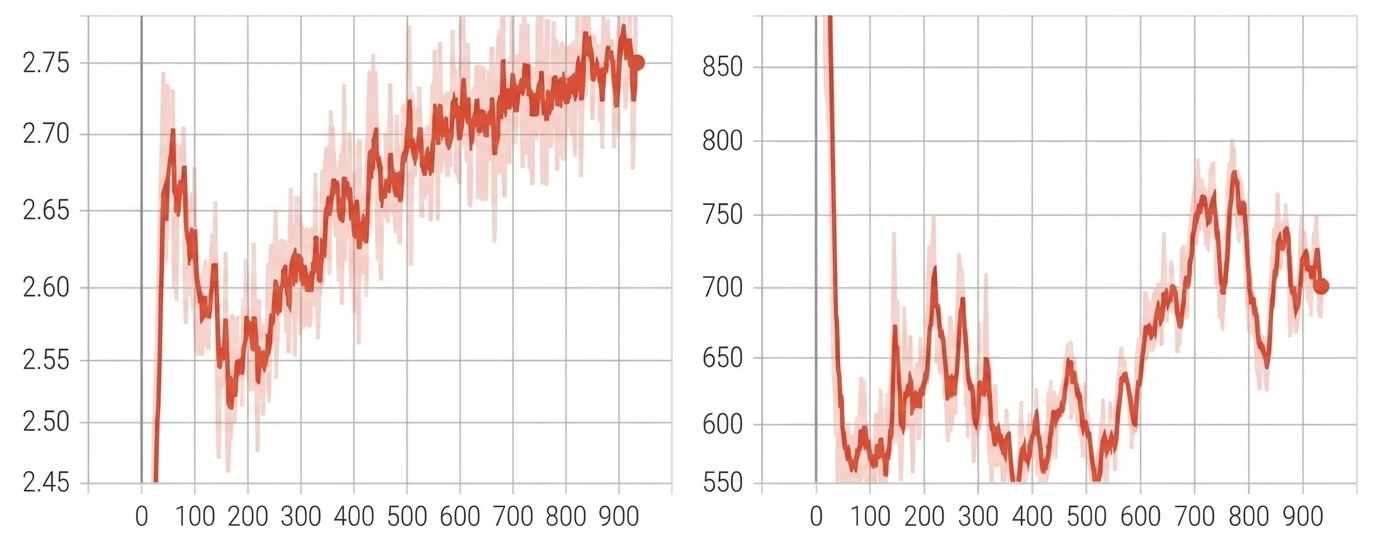}
\caption{Training curves of DCRL. Left: reward curve. Right: Mean response length per step.}
\label{fig:training_curve}
\end{figure}

The curriculum ablation further shows that progressive difficulty adjustment is useful. Uniformly sampled RL already outperforms easy-only or hard-only subsets, but the proposed dynamic curriculum delivers the best result, improving over uniformly sampling by +5.2 points. Figure~\ref{fig:training_curve} further shows stable convergence during DCRL training. Full ablation details remain in the supplement.

\section{Conclusion}

We introduced ReasonMatch-Bench, a benchmark for evaluating wide-baseline spatial correspondence in MLLMs, together with a scalable video-3D data pipeline and DCRL, a reinforcement-learning framework for training on this task with verifiable rewards. Across open- and closed-source baselines, current models remain substantially below human performance on a difficult 90-sample subset, where human annotators achieve 84.0 F1 and our best model reaches 52.0. Our experiments show that DCRL improves wide-baseline matching and yields positive transfer to several related spatial benchmarks while maintaining general visual understanding performance. These results suggest that wide-baseline correspondence is a useful testbed for studying cross-view spatial reasoning in MLLMs, and that substantial headroom remains.

\section*{Acknowledgments}
This work was supported in part by the Pioneer R\&D Program of Zhejiang (Grant No.~2025C01011), 
by the Ant Group Research Intern Program, and by 
and the National Natural Science Foundation of China (Grant No.~62576315).

{
    \small
    \bibliographystyle{ieeenat_fullname}

}

\clearpage
\setcounter{page}{1}
\maketitlesupplementary

\newtcolorbox{promptbox}{
    breakable,
    enhanced,
    colback=blue!5!white,         colframe=blue!40!black,       coltitle=white,               colbacktitle=blue!60!black,  fonttitle=\bfseries\large,   title=Matching Prompt,
    boxrule=1.0pt,                arc=6pt,                      top=6pt, bottom=6pt, left=8pt, right=8pt,
    attach boxed title to top center={yshift=-1.5mm},
    before skip=10pt, after skip=10pt,
    width=\textwidth,
    boxed title style={
            colframe=blue!70!black,
            colback=blue!70!black,
            sharp corners,
            boxrule=0pt,
            top=1pt,
            bottom=1pt,
            left=4pt,
            right=4pt,
        },
    drop shadow,
}

\section{Appendix Overview}

This supplementary material provides comprehensive details supporting our main paper, organized as follows:

\textbf{Sec.~\ref{sec:implement} -- Implementation Details:} Complete specifications of data generation pipeline, experimental setup, prompt template, and curriculum progression schedules.

\textbf{Sec.~\ref{sec:ablation} -- Additional Ablations:} Extended analyses on curriculum design variants, overlap scheduling strategies, and detailed RL vs. SFT comparisons across training stages.

\textbf{Sec.~\ref{sec:reasonmatch_results} -- Benchmark Analysis:} Detailed benchmark statistics, per-category model performance, failure mode analyses, and qualitative examples illustrating model capabilities and limitations.

\textbf{Sec.~\ref{sec:limitation} -- Extended Discussion:} Expanded analysis of limitations and concrete future research directions for advancing MLLM spatial intelligence.

\section{Implementation Details}
\label{sec:implement}

\subsection{Dataset Generation}

\noindent \textbf{Correspondence Generation from RGB-D Data.}
Starting from RGB-D videos (or RGB with known intrinsics and camera-to-world poses), we back-project every pixel $\mathbf{x}$ with valid depth in image $I_1$ into a 3D point $\mathbf{X}$ in the world coordinate system under a static-scene assumption, and reproject it to the consecutive image $I_2$ using the known camera parameters to obtain the image location $\mathbf{y}^{\text{proj}}$ and its camera depth $z^{\text{proj}}$. At $\mathbf{y}^{\text{proj}}$ we bilinearly sample the observed depth $z^{\text{obs}}$ and the color $I_2$.
We source our data from diverse datasets including uCO3D~\cite{liu24uco3d}, CO3D~\cite{reizenstein2021common}, and ScanNet~\cite{dai2017scannet}.
We compute two verifiable consistency terms:
(i) relative depth consistency
\begin{equation}
e_{\text{depth}}=\frac{\bigl|z^{\text{proj}} - z^{\text{obs}}\bigr|}{z^{\text{proj}}+\varepsilon},
\label{eq:depth}
\end{equation}
and (ii) photometric consistency defined as the channel-averaged RGB difference with bilinear sampling at $\mathbf{y}^{\text{proj}}$
\begin{equation}
e_{\text{photo}}=\tfrac{1}{3}\sum_{c\in\{r,g,b\}}\bigl|I_1^{(c)}(\mathbf{x})-\hat I_2^{(c)}(\mathbf{y}^{\text{proj}})\bigr|.
\label{eq:photo}
\end{equation}
A correspondence is considered valid if it meets basic visibility, boundedness criteria:
\begin{equation}
\mathbb{I}\bigl[\text{finite}(\mathbf{y}^{\text{proj}}) \wedge \text{in-bounds}(\mathbf{y}^{\text{proj}}) \wedge z^{\text{proj}}>0 \wedge z^{\text{obs}}>0\bigr],
\end{equation}
and error criteria:
\begin{equation}
\mathbb{I}\bigl[e_{\text{depth}}<\tau_{\text{d}}\bigr]\cdot
\mathbb{I}\bigl[e_{\text{photo}}<\tau_{\text{p}}\bigr].
\end{equation}
Here $\tau_{\text{d}}$ and $\tau_{\text{p}}$ are pre-defined thresholds for depth and photometric errors, respectively. Optional masks (e.g., dynamic-object or invalid-depth masks) can be applied on the image $I_1$ and $I_2$.

\noindent \textbf{Correspondence Generation from SfM Data.}
For datasets lacking ground-truth depth, such as RealEstate10k~\cite{zhou2018stereo} and DL3DV~\cite{ling2024dl3dv}, we leverage 3D reconstructions from Structure-from-Motion (SfM). For DL3DV, we utilize the provided COLMAP~\cite{schonberger2016structure} models directly. For RealEstate10k, which provides only RGB video streams, we first process the raw sequences with COLMAP to generate these 3D reconstructions.
Ground-truth correspondences are then extracted by identifying shared 3D landmarks between two image views. Specifically, a pair of 2D keypoints $(\mathbf{x}_i, \mathbf{y}_j)$, where $\mathbf{x}_i$ is in image $I_1$ and $\mathbf{y}_j$ is in image $I_2$, is considered a valid match if they both correspond to the same 3D point $\mathbf{X}$ in the COLMAP model's sparse reconstruction. This inlier set, having already passed geometric verification within COLMAP, serves as our reliable matches, bypassing the consistency checks (Eq.~\eqref{eq:depth} and \eqref{eq:photo}) used for RGB-D data.

\noindent \textbf{Quantifying Viewpoint Change.}
To categorize the difficulty of a pair, we use a scalar overlap score $\omega\!\in[0,1]$ as a proxy for co-visibility (higher means more similar views).
For pairs derived from RGB-D data, the overlap $\omega$ is defined as the proportion of validly matched pixel pairs $\mathcal{M}$ relative to the total number of pixels in each frame with height $H$ and width $W$:
\begin{equation}
\omega = \frac{|\mathcal{M}|}{H \times W}
\label{eq:overlap_rgbd}
\end{equation}
For pairs derived from SfM data, we define this overlap based on the proportion of shared 3D landmarks. Let $L_1$ and $L_2$ be the sets of 3D landmarks (i.e., 3D points) visible in images $I_1$ and $I_2$, respectively. The overlap $\omega$ is computed as:
\begin{equation}
\omega = \frac{|L_1 \cap L_2|}{\min(|L_1|, |L_2|)}
\label{eq:overlap_sfm}
\end{equation}
where $|\cdot|$ denotes the cardinality of the set.
The viewpoint-change magnitude is then defined as
\begin{equation}
\Delta_v \;=\; 1-\omega,
\label{eq:vchange}
\end{equation}
which is monotonically larger for more disparate viewpoints. In curriculum-based generation, stages specify an admissible overlap interval $[\underline{\omega}_s,\overline{\omega}_s]$, and a sample is routed to stage $s$ if
\begin{equation}
\mathbb{I}\bigl[\underline{\omega}_s \le \omega \le \overline{\omega}_s \bigr]=1.
\label{eq:stage_select}
\end{equation}

\noindent \textbf{Query Construction.}
Given a raw pair routed to stage $s$ via its overlap score (Eq.~\eqref{eq:stage_select}), we take its set of candidate matches $\mathcal{M}=\{(\mathbf{x}_k, \mathbf{y}_k)\}$ generated from either the RGB-D or SfM pipeline. We then select a spatially diverse \emph{core} subset $\mathcal{C}\subset\mathcal{M}$ under stage-specific sparsity constraints. By default, the tool uses a cluster-then-prune policy: matches are clustered in the joint space with DBSCAN at radius $\varepsilon=\alpha\cdot \tau_{\text{min-dist}}$ (with $\alpha\!\in(0,1)$), one representative per cluster is kept (closest to the cluster centroid), and a greedy max-spacing pass further reduces to at most $K$ points if needed.
Core matches are given a one-to-one mapping by sampling label subsets $\mathcal{L}_A^{\text{core}},\mathcal{L}_B^{\text{core}}$ of equal size, which satisfy the bijective constraint
\begin{equation}
f_{\text{core}}: \mathcal{L}_A^{\text{core}} \leftrightarrow \mathcal{L}_B^{\text{core}}.
\end{equation}
To increase difficulty, we can optionally add \emph{distractor} points $\mathcal{D}_A,\mathcal{D}_B$ sampled from the leftover matches in each view, ensuring no overlap with core matches.
Shuffling is applied to the labels in each view to avoid positional bias.
Finally, the benchmark sample is packaged as
\begin{equation}
(I_1, \mathcal{X}=\mathcal{C}_A \cup \mathcal{D}_A;\quad
I_2, \mathcal{Y}=\mathcal{C}_B \cup \mathcal{D}_B),
\end{equation}
with ground-truth mapping
\begin{equation}
f: \mathcal{L}_A \rightarrow \mathcal{L}_B \cup \{\varnothing\},
\end{equation}
where $f$ extends $f_{\text{core}}$ by assigning distractors to $\varnothing$.

\subsection{Dynamic Curriculum Strategy}

\paragraph{Verifiable Reward Design.}
Our reward function comprises two components that jointly encourage geometric accuracy and structured reasoning. Given the predicted mapping $\hat{f}: \{1,\ldots,n\} \rightarrow \{1,\ldots,m\} \cup \{\varnothing\}$ and ground-truth mapping $f^*$ with the same domain, where $n$ is the total number of query regions (including those with no correspondences in ground truth), we define:

\noindent\textbf{Format compliance} $r_{\text{format}} \in \{0, 1, 2\}$ verifies both structural and syntactic validity:
\begin{equation}
r_{\text{format}} = \mathbbm{1}[\text{structure}] + \mathbbm{1}[\text{JSON}],
\end{equation}
where the structure component checks that the output follows the required format with reasoning confined within thinking tags, and the JSON component verifies that the content within answer tags forms a valid JSON mapping with $\hat{f}(i) \in \{1,\ldots,m\} \cup \{\varnothing\}$ for all $i \in \{1,\ldots,n\}$.

\noindent\textbf{Matching correctness} $r_{\text{match}} \in [0, 1]$ measures prediction accuracy over all $n$ query regions:
\begin{equation}
r_{\text{match}} = \frac{1}{n}\sum_{i=1}^{n} \mathbbm{1}\left[\hat{f}(i) = f^*(i)\right],
\end{equation}
which counts exact agreements including correct predictions of unmatched regions ($\hat{f}(i) = \varnothing$ when $f^*(i) = \varnothing$). This formulation ensures that all regions are evaluated, rewarding comprehensive spatial reasoning.

The final reward combines these components:
\begin{equation}
r = w_f \cdot r_{\text{format}} + w_m \cdot r_{\text{match}},
\end{equation}
$w_f$ and $w_m$ being reward weights. Both components must be satisfied for high rewards, encouraging the model to produce well-structured and geometrically accurate predictions.
\subsection{Curriculum Learning Implementation Details}

Our curriculum design operates at two hierarchical levels: (1) three \textbf{cardinality settings} (L1, L2, L3) that define matching task configurations, and (2) three \textbf{training stages} that progressively combine these settings with increasing complexity.

\paragraph{Cardinality settings.}
We define three cardinality configurations corresponding to different matching scenarios:

\begin{itemize}
    \item \textbf{L1 (Unambiguous matching):} $n \sim \mathcal{U}(3, 5)$ matchable points with $|\mathcal{X}^{\text{dist}}| = |\mathcal{Y}^{\text{dist}}| = 0$. This yields 3--5 one-to-one correspondences without distractors.
    
    \item \textbf{L2 (Selective matching):} $n \sim \mathcal{U}(1, 2)$ matchable points with $|\mathcal{X}^{\text{dist}}| = 0$ and $|\mathcal{Y}^{\text{dist}}| \sim \mathcal{U}(3, 6)$. The model must identify 1--2 correct matches among 4--8 candidates in $\mathcal{Y}$.
    
    \item \textbf{L3 (Partial matching):} $n \sim \mathcal{U}(3, 6)$ matchable points with $|\mathcal{X}^{\text{dist}}|, |\mathcal{Y}^{\text{dist}}| \sim \mathcal{U}(3, 6)$. Both views contain 6--12 points with bidirectional occlusion.
\end{itemize}

\paragraph{Three-stage curriculum progression.}
Training progresses through three stages that combine cardinality settings with increasing complexity:

\begin{itemize}
    \item \textbf{Stage 1 (L1 only):} The model trains exclusively on unambiguous matching (L1) to establish foundational geometric reasoning. Stage transition occurs when the sliding window reward $\bar{r} > 0.7$ for 10 consecutive evaluations.
    
    \item \textbf{Stage 2 (L2 only):} After mastering L1, training shifts entirely to selective matching (L2) to learn distractor rejection. The model progresses when $\bar{r} > 0.7$ is maintained.
    
    \item \textbf{Stage 3 (L1/L2/L3 mixed):} The final stage samples from all three settings with probability $p_{\text{L1}} : p_{\text{L2}} : p_{\text{L3}} = 0.3 : 0.3 : 0.4$, ensuring comprehensive exposure to diverse matching scenarios.
\end{itemize}

\noindent\textbf{Adaptive demotion:} If performance degrades ($\bar{r}_{match} < 0.2$ for 10 consecutive steps) during any stage, the curriculum temporarily reverts to the previous stage for stabilization before resuming progression.

\paragraph{Spatial distribution refinement within settings.}
Independent of the curriculum stage, each cardinality setting (L1, L2, L3) undergoes its own spatial refinement process. This creates a two-dimensional curriculum: the \textit{stage} determines which cardinality settings are active, while \textit{spatial refinement} modulates difficulty within each active setting.

For each cardinality setting, we progressively eliminate local spatial cues:

\begin{itemize}
    \item \textbf{Initial sampling (clustered):} Points are sampled using DBSCAN clustering at radius $\varepsilon = \alpha \cdot \tau_{\text{min-dist}}$ followed by greedy max-spacing selection. Initially, $\alpha = 2.0$ and $\tau_{\text{min-dist}}$ equals the minimal non-overlap margin for annotations, producing spatially coherent clusters.
    
    \item \textbf{Progressive tightening:} Every time a level's internal reward threshold is satisfied, we update:
    \begin{align}
        \tau_{\text{min-dist}} &\leftarrow \max(\text{safe margin}, \tau_{\text{min-dist}} - 20) \text{ pixels}
    \end{align}
    This gradually decreases minimum point separation and reduces clustering radius.
    
    \item \textbf{Final sampling (dispersed):} When the final threshold is achieved inside a level, sampling transitions to pure greedy max-spacing, maximizing spatial dispersion and eliminating local context cues.
\end{itemize}

Importantly, spatial refinement operates \textit{independently} for each cardinality setting. When Stage 2 begins, L2 starts with clustered sampling even though L1 may have progressed to dispersed sampling. This ensures each matching configuration is learned from coarse to fine spatial distributions.

\subsection{Prompt Design}

We design a structured prompt that clearly specifies the cross-view matching task, input format, and expected output structure. As illustrated in Fig.~\ref{fig:prompt-design}, the prompt instructs the model to identify corresponding keypoint locations in a target image given marked query points in the source image. The prompt explicitly defines reasoning requirements, output format requirements (JSON structure with match validity flags). This design ensures consistent task interpretation across different models while enabling verifiable evaluation through structured output parsing.

\begin{figure*}[t]
    \centering
    {\small
        \begin{promptbox}
            \begin{itemize}[leftmargin=1em, itemsep=0pt, topsep=2pt]
            \item You are given two images of the same physical scene, each having several regions annotated with circles and IDs (e.g., "1", "2", "3").Your task is to identify the underlying correspondence between regions in the two images. Note that \textbf{maybe not} every region in the first image has a match in the second.Please provide your response in \textbf{two parts}: thinking process and final answer, each wrapped by some special tags.

            \item \texttt{thinking process}: Your analysis where you show your analyzing and thinking process wrapped in \texttt{<thinking> </thinking>} tags, which includes but not limited to:
            \begin{itemize}
                \item 1. \textbf{Describe Visual Regions}: For each annotated area in both images, describe the key visual characteristics and **spatial context within the scene**, including:
                    - **Intrinsic properties**: color, shape, texture, size, material
                    - **Spatial relationships in the 3D scene** (NOT pixel coordinates):
                        * What objects/structures are directly adjacent to this region? (e.g., "attached to a wooden door", "sitting on a metal shelf")
                        * What is this region positioned relative to in the physical space? (e.g., "below the window", "behind the chair", "left side of the bookshelf")
                        * Semantic context: What functional area or object group does it belong to? (e.g., "part of the dining area", "on the workspace desk")
                    - **Avoid** describing regions by their pixel locations (top-left, center-right, etc.) unless necessary for disambiguation
                    - Focus on **scene-level landmarks** as reference points (e.g., "near the entrance", "opposite to the main table", "in the corner with the lamp")
                \item 2. \textbf{Compare Viewpoints}: Analyze the geometric or perspective relationship between the two images, such as:
                    - Overview of the two images' contents, focusing on **how the physical scene layout appears** in each view
                    - **Camera transformation**: rotation angle (e.g., "camera rotated ~90° clockwise around the room center"), translation (e.g., "camera moved closer to the left wall"), zoom/scale differences
                    - **Occlusion changes**: which objects/regions become visible or hidden due to viewpoint change?
                    - **Perspective distortion**: how do spatial relationships appear to change due to different viewing angles? (e.g., "objects on the right side now appear more frontal")
                \item 3. \textbf{Infer Matching Relations}: Based on region appearance and **scene-relative spatial relationships**, establish correspondences:
                    - Iterate through each annotated region in image 1, and for each one, compare it sequentially with every annotated region in image 2 to infer the matching likelihood and reasonableness of region correspondences between the two images
                    - Prioritize matching based on **what surrounds each region** and **3D spatial context** rather than 2D image positions
                    - Use **stable scene anchors** (walls, large furniture, architectural features) to reason about region identity across views
                    - Consider how the viewpoint change transforms the **spatial relationships** you identified in step 1
                    - Example reasoning: "Region A-1 is next to a red door and below a window. Region B-3 is also adjacent to the same red door (now seen from a different angle) and below the same window structure, so A-1 matches B-3"
            \end{itemize}
            \item \texttt{final answer}: Your final answer based on your thinking part wrapped by \texttt{<answer> </answer>} tags. The JSON object is a mapping of region IDs from image A (as string keys) to the corresponding region IDs in image B (as string values). **For regions in A that have no match in B, use "none" as the value.**

\end{itemize}
        \end{promptbox}
    }
    \caption{Prompt For Wide-Baseline Matching Task.
    }
    \label{fig:prompt-design}
\end{figure*}

\section{Ablation Studies}
\label{sec:ablation}

\paragraph{Reinforcement Learning vs. Supervised Fine-tuning.}
To assess our curriculum-based reinforcement learning approach, we compare against supervised fine-tuning (SFT) on the same cross-view matching data. The SFT baseline uses 300 steps of supervised training with teacher-forced matching predictions, while our DCRL method employs reinforcement learning with holistic geometric rewards as described in Sec.~\ref{sec:method}.

\textbf{General vision-language understanding.} As shown in Table~\ref{tab:ablation_general}, DCRL maintains strong performance on general vision-language benchmarks and modestly improves over both the base model and the SFT baseline. While SFT degrades on MMStar ($-$3.4 points) and V* ($-$2.6 points), potentially due to catastrophic forgetting under domain shift, our RL-based approach preserves and slightly improves general capabilities ($+$2.7 on MMStar, $+$1.1 on V*). This result suggests that curriculum-based reinforcement learning with geometric rewards can strengthen spatial training without harming pre-existing vision-language understanding.

\begin{table}[t]
\centering

\caption{Ablation study on general vision-language understanding benchmarks. Both SFT and DCRL maintain strong performance on general tasks, with DCRL showing slight advantages in our evaluation.}
\label{tab:ablation_general}
\small
\setlength{\tabcolsep}{6pt}
\resizebox{\linewidth}{!}{
\begin{tabular}{l cccc}
\toprule
\textbf{Method} & \textbf{MME-RealWorld} & \textbf{MMStar} & \textbf{RealWorldQA} & \textbf{V*} \\
\midrule
Base Model & 62.8 & 59.8 & 69.5 & 84.8 \\
SFT & 62.4 & 56.4 & 68.8 & 82.2 \\
\midrule
\textbf{Ours (DCRL)} & \textbf{63.8} & \textbf{62.5} & \textbf{70.5} & \textbf{85.9} \\
\midrule
\end{tabular}
}
\end{table}

\textbf{Spatial intelligence and geometric reasoning.} Table~\ref{tab:ablation_spatial} shows notable advantages of RL over SFT on spatial reasoning tasks. On our cross-view matching benchmark (ReasonMatch), DCRL achieves 70.5\% compared with 51.0\% for SFT, a 19.5-point improvement. The contrast is even larger on SAT (+34.0 points), where SFT drops from the base model (70.0 $\to$ 41.3) while RL improves over it (70.0 $\to$ 75.3). This difference is consistent with the view that teacher-forced imitation of specific matching patterns may be less robust when task distributions differ.
\begin{table}[tb]
\centering
\caption{Ablation study on spatial intelligence and geometric reasoning benchmarks. DCRL substantially outperforms SFT, particularly on tasks requiring fine-grained geometric correspondence (SAT, ReasonMatch), which is consistent with a benefit from curriculum-based reinforcement learning for spatial reasoning.}
\label{tab:ablation_spatial}
\small
\setlength{\tabcolsep}{6pt}
\resizebox{\linewidth}{!}{
\begin{tabular}{l cccc}
\toprule
\textbf{Method} & \textbf{OmniSpatial} & \textbf{MindCube} & \textbf{SAT} & \textbf{ReasonMatch} \\
\midrule
Base Model & 43.6 & 40.0 & 70.0 & 27.5 \\
SFT & 42.6 & 45.1 & 41.3 & 51.0 \\
\midrule
\textbf{Ours (DCRL)} & \textbf{48.9} & \textbf{43.5} & \textbf{75.3} & \textbf{70.5} \\
\bottomrule
\end{tabular}
}
\end{table}

One possible explanation for the performance contrast is the difference in training dynamics. SFT encourages the model to imitate exact correspondence patterns from training data, which may create rigid associations that generalize poorly and can override useful pre-existing knowledge. In contrast, RL permits exploration of diverse matching strategies guided by holistic geometric feedback, which may help preserve prior capabilities while improving the target task. The dynamic curriculum may further support this process by progressively exposing harder scenarios rather than relying on a single fixed difficulty level.

Across spatial benchmarks, DCRL improves over the base model on all four reported tasks ($+$43.0 on ReasonMatch, $+$5.3 on SAT, $+$5.3 on OmniSpatial, and $+$3.5 on MindCube). By contrast, SFT shows mixed behavior: it improves on ReasonMatch ($+$23.5) and MindCube ($+$5.1), is roughly flat on OmniSpatial ($-$1.0), and degrades substantially on SAT ($-$28.7). This pattern suggests that supervised imitation may be less robust for transferable spatial reasoning in this setting, while our RL approach produces gains that transfer more consistently across the reported geometric benchmarks.

\begin{table}[t]
\centering
\caption{Ablation study on curriculum learning. Progressive difficulty adjustment across viewpoint divergence and correspondence complexity improves performance over uniform sampling and fixed-difficulty training.}
\label{tab:ablation_curriculum}
\small
\setlength{\tabcolsep}{8pt}
\resizebox{\linewidth}{!}{
\begin{tabular}{l c}
\toprule
\textbf{Training Strategy} & \textbf{ReasonMatch} \\
\midrule
No Curriculum (uniform cardinality) & 65.3 \\
Only 1/4 minimal $\Delta_v$ samples (Easy) & 59.9 \\
Only 1/4 maximum $\Delta_v$ samples (Hard) & 62.3 \\
\midrule
\textbf{Dynamic Curriculum (Ours)} & \textbf{70.5} \\
\midrule
$\Delta$ (Ours vs. No Curriculum) & +5.2 \\
$\Delta$ (Ours vs. Easy) & +10.6 \\
$\Delta$ (Ours vs. Hard) & +8.2 \\
\bottomrule
\end{tabular}
}
\end{table}

\paragraph{Curriculum Learning.}
To assess the importance of our dynamic three-dimensional curriculum, we compare against three ablated training strategies: (1) uniform sampling without curriculum, (2) training only on easy samples (first quartile by viewpoint overlap), and (3) training only on hard samples (last quartile with minimal overlap). All variants use identical RL training setup and total training steps.

As shown in Table~\ref{tab:ablation_curriculum}, our dynamic curriculum achieves 70.5\% F1, outperforming uniform sampling (65.3\%, +5.2 points). This result suggests that progressively increasing viewpoint divergence, correspondence complexity, and spatial distribution challenges can be more effective than treating all samples equally. A plausible explanation is that the curriculum allows the model to build useful intermediate competence on simpler configurations before tackling extreme viewpoint changes.

Training exclusively on easy or hard samples yields significantly worse results. The easy-only strategy (59.9\%) underperforms uniform sampling by 5.4 points, suggesting that limiting training to small viewpoint changes and simple correspondence structures is insufficient in this setup. Conversely, training only on hard samples (62.3\%) also underperforms uniform sampling (-3.0 points), suggesting that starting with extreme viewpoint divergence and complex correspondences is also less effective than gradual difficulty scheduling.

The substantial gaps between curriculum and single-difficulty training (+10.6 over easy-only, +8.2 over hard-only) are consistent with the value of progressive learning. Our curriculum's adaptive adjustment mechanism monitors sustained performance improvements before increasing difficulty, helping the model stabilize before moving to harder settings. This staged progression across three dimensions (viewpoint span, correspondence complexity, and spatial distribution) is also consistent with the transfer gains observed on OmniSpatial, MindCube, and SAT.

Interestingly, hard-only training (62.3\%) slightly outperforms easy-only (59.9\%), suggesting that exposure to challenging cases, even without curriculum structure, may provide more useful learning signals than trivial examples. However, both approaches remain well below curriculum learning, indicating that progressive difficulty scheduling is beneficial in this setting.

\begin{table*}[ht]
\centering
\caption{Human study on the 90 largest view divergence samples across three datasets. Human annotators achieve substantially higher performance than all tested models, with the performance gap most pronounced on object-centric scenes (uCO3D).}
\label{tab:human_study}
\small
\setlength{\tabcolsep}{3pt}
\begin{tabular}{l ccc ccc ccc ccc}
\toprule
\textbf{Method} & 
\multicolumn{3}{c}{\textbf{Overall}} & 
\multicolumn{3}{c}{\textbf{DL3DV}} & 
\multicolumn{3}{c}{\textbf{RE10K}} & 
\multicolumn{3}{c}{\textbf{uCO3D}} \\
\cmidrule(r){2-4} \cmidrule(lr){5-7} \cmidrule(lr){8-10} \cmidrule(l){11-13}
& \textbf{P} & \textbf{R} & \textbf{F1} & 
\textbf{P} & \textbf{R} & \textbf{F1} & 
\textbf{P} & \textbf{R} & \textbf{F1} & 
\textbf{P} & \textbf{R} & \textbf{F1} \\
\midrule
\multicolumn{13}{l}{\textit{Closed-Source Models}} \\
GPT-5-mini & 34.5 & 40.7 & 37.2 & 32.5 & 40.4 & 35.9 & 47.4 & 52.7 & 49.7 & 23.6 & 29.1 & 25.8 \\
Gemini-2.5-Pro & 28.3 & 31.1 & 29.5 & 25.1 & 28.3 & 26.5 & 42.8 & 45.6 & 44.1 & 17.1 & 19.3 & 18.0 \\
Claude-4.5-Sonnet & 22.1 & 26.7 & 24.0 & 20.6 & 24.8 & 22.2 & 28.1 & 33.6 & 30.5 & 17.4 & 21.6 & 19.2 \\
\midrule
\multicolumn{13}{l}{\textit{Open-Source Models}} \\
Qwen3-VL-235B & 27.5 & 33.0 & 29.9 & 23.2 & 28.1 & 25.3 & 42.3 & 50.2 & 45.7 & 17.1 & 20.9 & 18.7 \\
\midrule
\textbf{Ours (DCRL)} & \textbf{48.3} & \textbf{56.8} & \textbf{52.0} & \textbf{53.3} & \textbf{63.2} & \textbf{57.7} & \textbf{65.7} & \textbf{76.6} & \textbf{70.6} & \textbf{25.9} & \textbf{30.4} & \textbf{27.8} \\
\midrule
\rowcolor{gray!15}
\textbf{Human} & \textbf{87.5} & \textbf{81.3} & \textbf{84.0} & \textbf{94.3} & \textbf{93.3} & \textbf{93.5} & \textbf{96.1} & \textbf{93.9} & \textbf{94.7} & \textbf{72.2} & \textbf{56.6} & \textbf{62.1} \\
\midrule
$\Delta$ (Human vs. Ours) & +39.2 & +24.5 & +32.0 & +41.0 & +30.1 & +35.8 & +30.4 & +17.3 & +24.1 & +46.3 & +26.2 & +34.3 \\
\bottomrule
\end{tabular}
\end{table*}

\section{ReasonMatch-Bench Results}
\label{sec:reasonmatch_results}

In this section, we report detailed results on ReasonMatch-Bench and analyze how different multimodal LLMs behave under our cross-view region matching task. We first summarize the overall quantitative performance across scenarios and difficulty levels and then provide a fine-grained comparison along several error dimensions where models exhibit the largest discrepancies. Finally, we briefly discuss a visualization that highlights these differences.

\subsection{Overall Quantitative Performance}
\label{subsec:sup_reasonmatch_overall}

Table~\ref{tab:model_comparison_percent} reports F1, Precision, and Recall on ReasonMatch-Bench for a broad range of multimodal LLMs, together with per-scenario scores on three difficulty levels (Indoor/Outdoor/Object, L1--L3).

\paragraph{ReasonMatch-Bench is challenging for strong general-purpose MLLMs.}
Even the best general-purpose models only achieve F1 scores in the 50--60 range on our benchmark.
For example, GPT-5-mini obtains an overall F1 of 57.9, while GPT-5-Chat and Qwen3-VL-235B reach 51.5 and 49.2, respectively.
Other strong closed-source systems such as Gemini-2.5-Pro, Claude-4.5-Sonnet, and GPT-4o remain in the 33--43 F1 range.
This indicates that ReasonMatch-Bench is substantially harder than standard VQA-style benchmarks, even for large commercial MLLMs.

\paragraph{DCRL turns a weak open-weight model into the best performer.}
The baseline performs poorly on ReasonMatch-Bench (27.5 \%), substantially below both commercial models and the larger Qwen3-VL-235B (49.2 \%).
After applying our DCRL training, the same 8B backbone achieves 70.5\% F1, outperforming all other models in the table by a large margin.
This suggests that targeted training on ReasonMatch-Bench-style tasks can, in this setting, be more effective than simply scaling up model size.

\paragraph{Performance consistently drops with difficulty, especially on Object L3.}
Across models, scores tend to decrease from L1 to L3 within each scenario, indicating that our difficulty annotation captures genuine changes in task hardness.
For instance, GPT-5-mini drops from 68.2 to 47.0 on Indoor (L1$\rightarrow$L3) and from 75.8 to 51.4 on Outdoor, and similar declines appear for other models.
The Object L3 column is consistently the hardest setting, and our Qwen3-VL+DCRL model still struggles on this regime (33.7) compared to easier levels, but remains clearly above all baselines.

\subsection{Fine-Grained Error Analysis Across Models}
\label{subsec:sup_reasonmatch_error}

\begin{figure}[ht]
    \centering
    \includegraphics[width=0.95\linewidth]{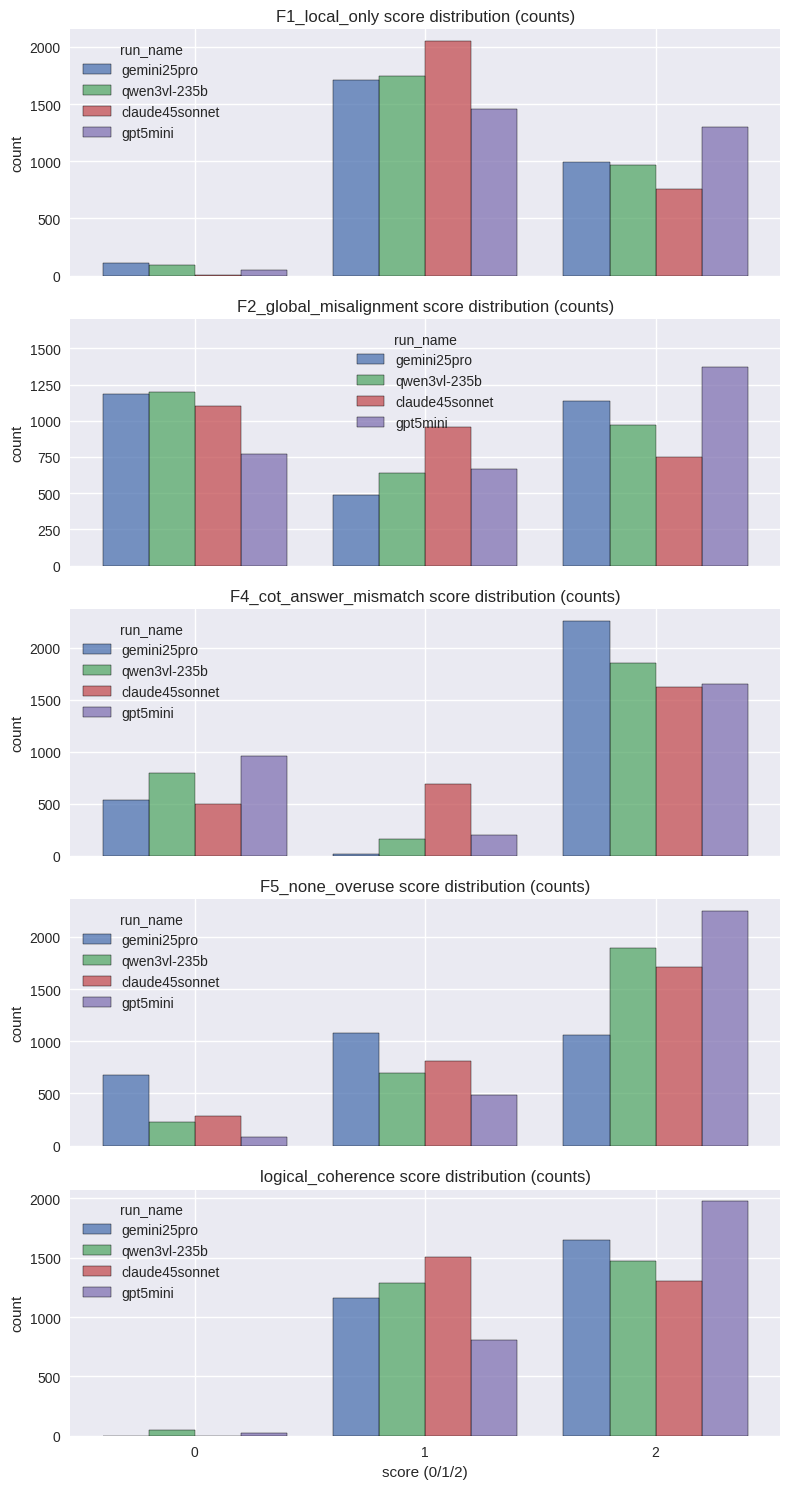}
    \caption{Comparison of five models across key error dimensions:
    Local-Cue Reliance (F1), Global Layout Misalignment (F2),
    Reasoning-Answer Mismatch (F4), Overuse of ``None'' (F5),
    and Reasoning Coherence.
    Higher bars indicate fewer or milder errors.}
    \label{fig:sup_reasonmatch_error_vis}
\end{figure}

Although Table~\ref{tab:model_comparison_percent} reflects how often models select the correct match, it does not explain \emph{why} they fail. To better characterize model behavior, we use Qwen3-VL-235B as a blind evaluator to score the complete ReasonMatch-Bench outputs of the five models shown in Figure~\ref{fig:sup_reasonmatch_error_vis}. For each example, the evaluator receives the two annotated images, the ground-truth mapping, the verbatim reasoning trace, and the parsed final answer, and assigns discrete scores in $\{0,1,2\}$ to a broader rubric, where 2 indicates no issue and 0 indicates a severe issue. Below, we report the five dimensions with the clearest cross-model differences: Local-Cue Reliance (F1), Global Layout Misalignment (F2), Reasoning-Answer Mismatch (F4), Overuse of ``None'' (F5), and Reasoning Coherence.

\paragraph{Local-Cue Reliance (F1).}
This score measures how well a model can integrate correct local observations into a globally consistent cross-view mapping. Low F1 reflects the typical failure pattern where the chain-of-thought describes the correct shelf, tier, or surrounding objects, but the final correspondence still points to a wrong region. Models with higher F1 scores are better at binding fine-grained visual cues into a coherent global decision.

\paragraph{Global Layout Misalignment (F2).}
F2 captures large-scale geometric errors such as swapped left/right ordering, incorrect vertical tiers, or inconsistent depth relations between regions. This dimension directly measures a model’s ability to maintain a viewpoint-consistent global layout rather than reasoning in a purely local, patch-based manner.

\paragraph{Reasoning-Answer Mismatch (F4).}
F4 quantifies inconsistencies between the model’s chain-of-thought and its final structured output. A low score indicates cases where the reasoning text identifies the correct target region, but the final JSON or \texttt{<answer>} block encodes a different ID or swaps indices. This dimension probes how reliably a model can translate its internal reasoning into a precise, machine-usable prediction.

\paragraph{Overuse of ``None'' (F5).}
ReasonMatch-Bench includes regions that legitimately have no correspondence in the other view. F5 measures whether the model uses the ``none'' option in a calibrated way. Over-using ``none''---especially on regions that do have correct matches---reveals an overly conservative behavior: the model can often describe the local content but refuses to commit to a specific correspondence.

\paragraph{Reasoning Coherence.}
Finally, we rate the overall logical coherence of the chain-of-thought: whether observations are ordered in a reasonable way, whether there are self-contradictions, and whether the final decision is clearly connected to previous observations. This dimension complements F1/F2 by highlighting cases where the answer is correct but the reasoning is disorganized, as well as cases where an apparently detailed reasoning trail actually does not support the final mapping.

\vspace{0.3em}
\noindent

Figure~\ref{fig:sup_reasonmatch_error_vis} summarizes these five dimensions across all evaluated models. We report mean rubric scores in $[0, 2]$, so that higher values consistently correspond to better behavior (fewer or milder errors). Several patterns emerge:

\begin{itemize}
    \item Models differ most strongly on \textbf{F1} and \textbf{F2}, suggesting that binding local cues into a globally consistent spatial alignment is the central bottleneck on ReasonMatch-Bench.
    \item \textbf{F4} reveals that some models frequently ``think correctly but answer incorrectly'', exposing weaknesses in reasoning-to-action alignment that are invisible to raw F1 / Precision / Recall.
    \item Differences in \textbf{F5} highlight distinct calibration behaviors: some models are overly cautious and tend to abstain by predicting ``none'', while others are more willing to commit to concrete correspondences.
    \item The \textbf{logical coherence} scores show that even when overall accuracy is similar, models can vary substantially in how structured and self-consistent their reasoning is.
\end{itemize}

Overall, these error dimensions provide a complementary perspective to the aggregate metrics in Table~\ref{tab:model_comparison_percent}. They show that current multimodal LLMs are limited not only by perception, but also by their ability to maintain a globally consistent 3D picture of the scene, propagate that picture through a coherent reasoning process, and translate it into a precise region-level correspondence.

\paragraph{Qualitative examples.}
To make the above error dimensions more concrete, we highlight a representative case for \texttt{F2} from ReasonMatch-Bench.
In a ScanNet indoor scene (\texttt{scannet\_000751}), several floor markers form a small cluster that appears in both views under essentially the same physical configuration.
The ground truth correspondence in Image~B is region~B-1, which is the leftmost marker in this cluster.
However, the model often selects region~B-3 and describes it as the leftmost marker.
This behavior indicates that the model has roughly identified the correct cluster, but fails to maintain a consistent left--right ordering when projecting the 3D layout into the two viewpoints, which is precisely what \texttt{F2} is designed to capture.

\subsection{Visualization of Model Behaviors}
\label{subsec:reasonmatch_vis}

To make the above differences more concrete, Figure~\ref{fig:sup_reasonmatch_error_vis} provides a compact visual summary of model behaviors along five error dimensions: Local-Cue Reliance (F1), Global Layout Misalignment (F2), Reasoning-Answer Mismatch (F4), Overuse of ``None'' (F5), and Reasoning Coherence.

This visualization highlights which aspects of spatial reasoning each model handles well and where it fails:
\begin{itemize}
    \item Models with high scores on F1 and F2 are able to build a coherent 3D understanding from two wide-baseline views, rather than relying on local texture cues alone.
    \item Low F4 and logical coherence expose brittle reasoning pipelines, where correct evidence is not reliably turned into the correct structured prediction.
    \item Variations in F5 reveal different risk profiles: some models trade recall for precision by overusing ``none'', while others achieve better coverage at the cost of more misaligned matches.
\end{itemize}

In combination with the quantitative scores in Table~\ref{tab:model_comparison_percent}, this figure shows that ReasonMatch-Bench is not just a harder matching dataset, but also a targeted probe of multi-view spatial understanding, calibration, and reasoning quality in multimodal LLMs.

\begin{figure*}[t]
    \centering
    \includegraphics[page=1,width=\textwidth]{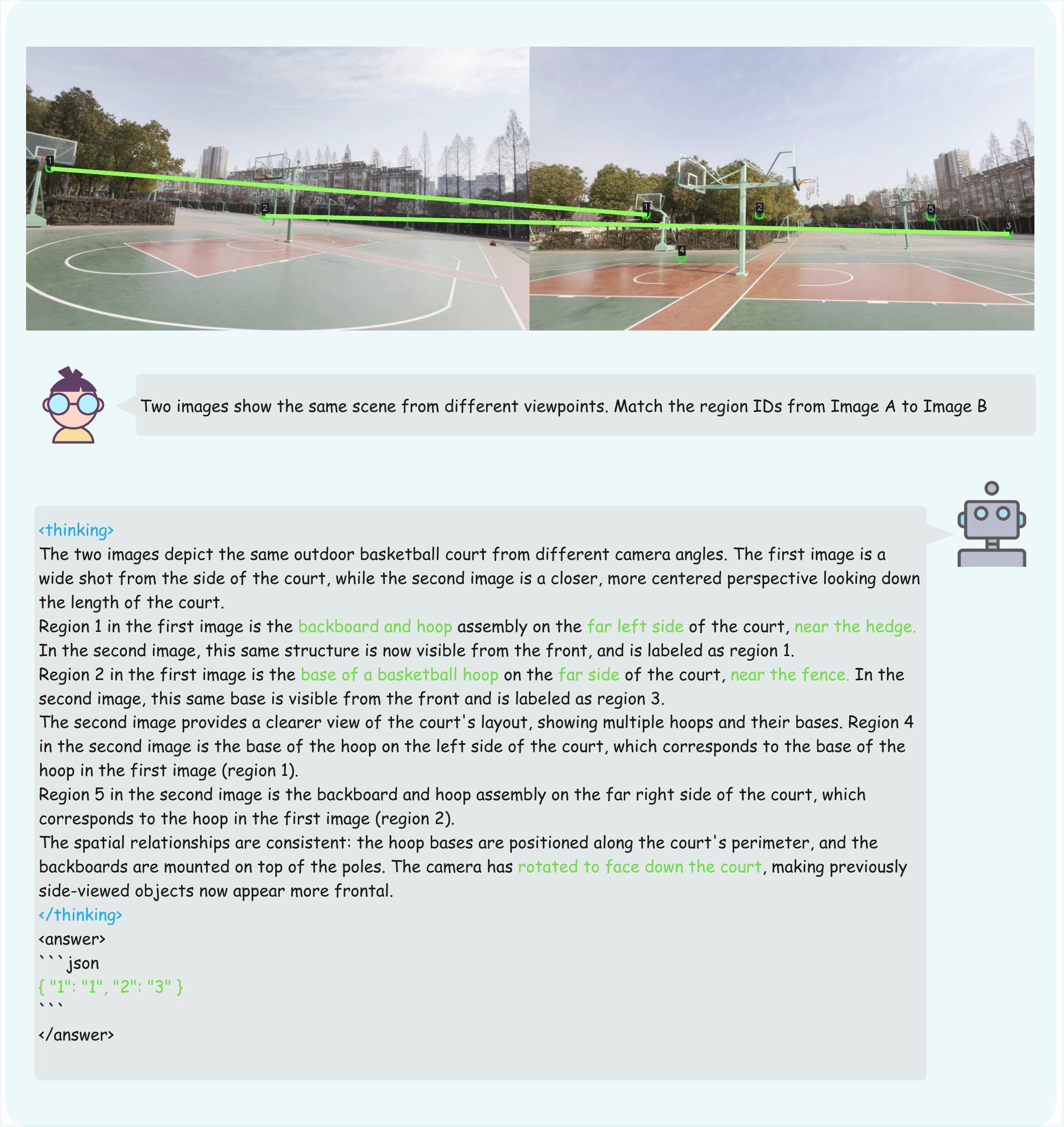}
    \caption{Qualitative example on ReasonMatch-Bench (Example~1). 
    The model is given two views of the same scene and asked to match region IDs from Image~A to Image~B.
    Its chain-of-thought correctly identifies the corresponding regions across viewpoints
    and produces a consistent JSON mapping for the cross-view region matching task.}
    \label{fig:reasonmatch_qual_success_1}
\end{figure*}

\begin{figure*}[t]
    \centering
    \includegraphics[page=2,width=\textwidth]{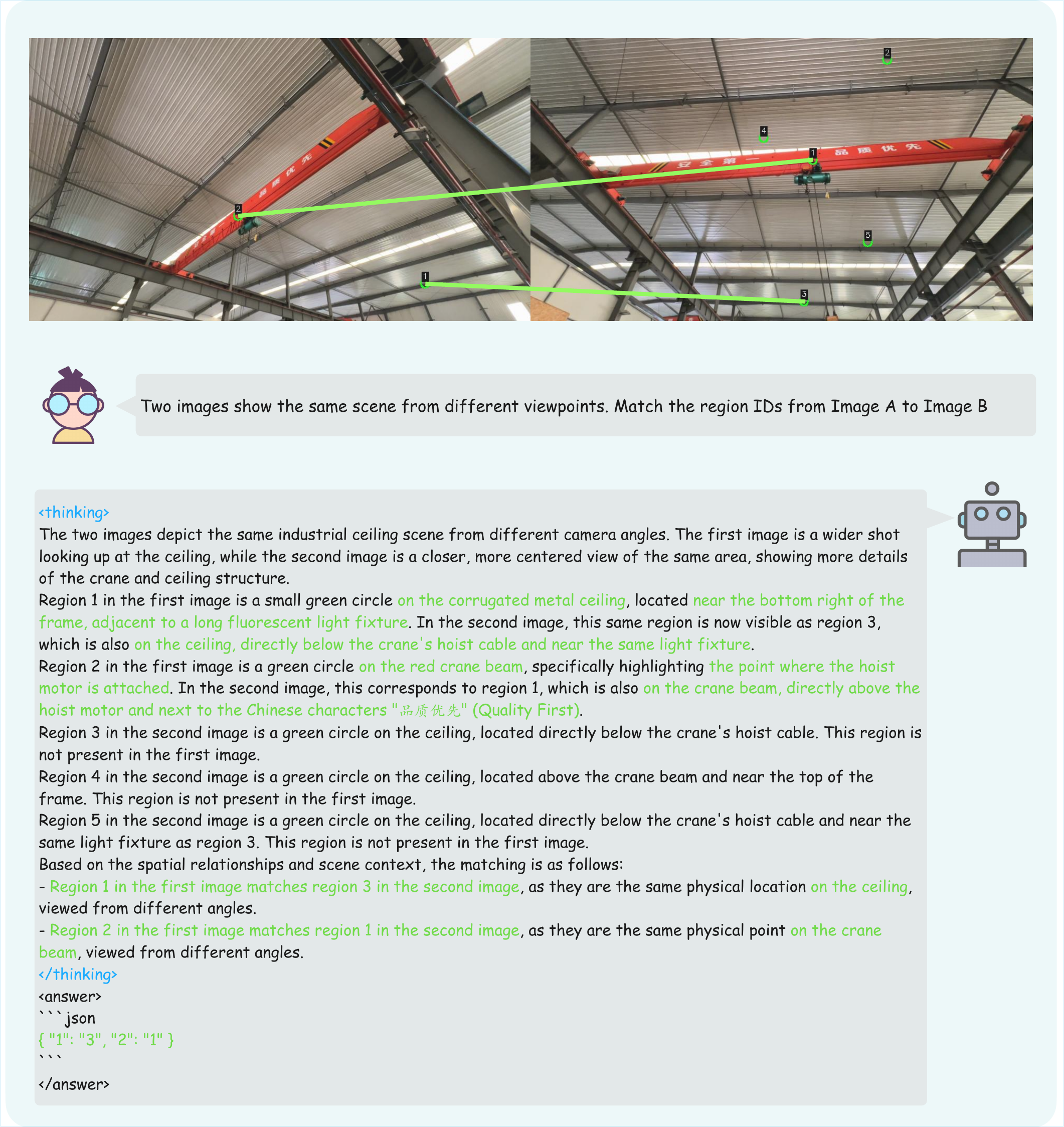}
    \caption{Qualitative example on ReasonMatch-Bench (Example~2). 
    Another successful case where the model aligns regions across wide-baseline views,
    using textual reasoning over geometric layout and shared objects to produce a correct JSON correspondence.
    Together with Figure~\ref{fig:sup_reasonmatch_error_vis}, these examples illustrate how multimodal LLMs
    solve our pixel-anchored cross-view matching task when their spatial reasoning is reliable.}
    \label{fig:reasonmatch_qual_success_2}
\end{figure*}

To further illustrate how models behave on individual instances, Figures~\ref{fig:reasonmatch_qual_success_1} and~\ref{fig:reasonmatch_qual_success_2} show full chain-of-thought traces and JSON outputs on two representative successful examples from ReasonMatch-Bench.

\subsection{Human Study}
\label{subsec:human_study}

To assess how the task aligns with natural human spatial reasoning and to measure the gap between human performance and current models, we conduct a human study on a stratified 90-sample subset with the largest view divergence. Two non-expert annotators, i.e., ordinary participants without task-specific training, complete the task independently using the same instructions as the models. They do not discuss their answers during annotation, and we report the average of their scores as the final human result.

Despite not having specialized training, the human annotators achieve an overall F1 of 84.0\%, substantially outperforming all tested models, including frontier systems (Table~\ref{tab:human_study}). The 32-point gap between the non-expert annotators (84.0\%) and our best model (52.0\%) highlights a large remaining challenge in fine-grained geometric correspondence.

Performance patterns differ markedly across scene types. On structured indoor and outdoor environments, humans achieve near-perfect accuracy (F1 $>$ 93\%), likely aided by rich contextual cues and stronger geometric regularity. Both humans and models struggle more on object-centric scenes (uCO3D), where humans achieve 62.1\% F1 compared with 27.8\% for our best model. Analysis of human annotations suggests that errors primarily occur when objects exhibit severe self-occlusion or lack distinctive surface features, which are also challenging conditions for current models. This pattern is consistent with the view that the benchmark captures genuine spatial reasoning challenges rather than artifacts of imperfect ground truth.

The human-model gap underscores substantial limitations in current MLLMs' spatial reasoning. While DCRL improves over strong baselines (+14.8\% over GPT-5-mini) and reaches 62\% of non-expert human performance overall, there remains considerable room before models approach human-level geometric reasoning. The fact that non-expert annotators can solve many of these tasks reliably further motivates the benchmark and continued work on spatial understanding in MLLMs.

\section{Limitations and Future Work}
\label{sec:limitation}

Despite substantial improvements over baselines, our approach reveals both the promise and challenges of developing human-level spatial intelligence in MLLMs. Our best model achieves 52.0\% F1 compared to untrained humans' 84.0\%, with the gap most pronounced on object-centric scenes where even humans struggle (27.8\% vs. 62.1\%). This persistent performance gap highlights fundamental limitations in current architectures' ability to perform fine-grained geometric reasoning under extreme viewpoint changes and limited contextual cues.

Our work focuses on pairwise cross-view matching as a foundational capability, but comprehensive spatial intelligence demands reasoning over multiple views simultaneously, integrating geometric correspondence with 3D scene understanding, temporal dynamics, and semantic knowledge. Future research should extend beyond pairwise matching toward holistic multi-view reasoning that mirrors human spatial cognition—synthesizing information across viewpoints to construct coherent 3D mental models.


\begin{thebibliography}{66}
\providecommand{\natexlab}[1]{#1}
\providecommand{\url}[1]{\texttt{#1}}
\expandafter\ifx\csname urlstyle\endcsname\relax
  \providecommand{\doi}[1]{doi: #1}\else
  \providecommand{\doi}{doi: \begingroup \urlstyle{rm}\Url}\fi

\bibitem[Bai et~al.(2025{\natexlab{a}})Bai, Cai, Chen, Chen, Chen, Cheng, Deng,
  Ding, Gao, Ge, et~al.]{bai2025qwen3}
Shuai Bai, Yuxuan Cai, Ruizhe Chen, Keqin Chen, Xionghui Chen, Zesen Cheng,
  Lianghao Deng, Wei Ding, Chang Gao, Chunjiang Ge, et~al.
\newblock Qwen3-vl technical report.
\newblock \emph{arXiv preprint arXiv:2511.21631}, 2025{\natexlab{a}}.

\bibitem[Bai et~al.(2025{\natexlab{b}})Bai, Chen, Liu, Wang, Ge, Song, Dang,
  Wang, Wang, Tang, Zhong, Zhu, Yang, Li, Wan, Wang, Ding, Fu, Xu, Ye, Zhang,
  Xie, Cheng, Zhang, Yang, Xu, and Lin]{bai2025qwen25vltechnicalreport}
Shuai Bai, Keqin Chen, Xuejing Liu, Jialin Wang, Wenbin Ge, Sibo Song, Kai
  Dang, Peng Wang, Shijie Wang, Jun Tang, Humen Zhong, Yuanzhi Zhu, Mingkun
  Yang, Zhaohai Li, Jianqiang Wan, Pengfei Wang, Wei Ding, Zheren Fu, Yiheng
  Xu, Jiabo Ye, Xi Zhang, Tianbao Xie, Zesen Cheng, Hang Zhang, Zhibo Yang,
  Haiyang Xu, and Junyang Lin.
\newblock Qwen2.5-vl technical report, 2025{\natexlab{b}}.

\bibitem[Banerjee et~al.(2025)Banerjee, Shkodrani, Moulon, Hampali, Han, Zhang,
  Zhang, Fountain, Miller, Basol, et~al.]{banerjee2025hot3d}
Prithviraj Banerjee, Sindi Shkodrani, Pierre Moulon, Shreyas Hampali, Shangchen
  Han, Fan Zhang, Linguang Zhang, Jade Fountain, Edward Miller, Selen Basol,
  et~al.
\newblock Hot3d: Hand and object tracking in 3d from egocentric multi-view
  videos.
\newblock In \emph{CVPR}, pages 7061--7071, 2025.

\bibitem[Barath and Matas(2018)]{barath2018graph}
Daniel Barath and Jiri Matas.
\newblock Graph-cut ransac.
\newblock In \emph{Proceedings of the IEEE conference on computer vision and
  pattern recognition}, pages 6733--6741, 2018.

\bibitem[Barath et~al.(2019)Barath, Matas, and Noskova]{barath2019magsac}
Daniel Barath, Jiri Matas, and Jana Noskova.
\newblock Magsac: marginalizing sample consensus.
\newblock In \emph{Proceedings of the IEEE/CVF conference on computer vision
  and pattern recognition}, pages 10197--10205, 2019.

\bibitem[Bay et~al.(2006)Bay, Tuytelaars, and Van~Gool]{bay2006surf}
Herbert Bay, Tinne Tuytelaars, and Luc Van~Gool.
\newblock Surf: Speeded up robust features.
\newblock In \emph{European conference on computer vision}, pages 404--417.
  Springer, 2006.

\bibitem[Bochkovskii et~al.(2024)Bochkovskii, Delaunoy, Germain, Santos, Zhou,
  Richter, and Koltun]{bochkovskii2024depth}
Aleksei Bochkovskii, Ama{\~A}{\c{G}}l Delaunoy, Hugo Germain, Marcel Santos,
  Yichao Zhou, Stephan~R Richter, and Vladlen Koltun.
\newblock Depth pro: Sharp monocular metric depth in less than a second.
\newblock \emph{arXiv preprint arXiv:2410.02073}, 2024.

\bibitem[Burkard et~al.(2012)Burkard, Dell'Amico, and
  Martello]{burkard2012assignment}
Rainer Burkard, Mauro Dell'Amico, and Silvano Martello.
\newblock \emph{Assignment problems: revised reprint}.
\newblock SIAM, 2012.

\bibitem[Chen et~al.(2024{\natexlab{a}})Chen, Xu, Kirmani, Ichter, Driess,
  Florence, Sadigh, Guibas, and Xia]{chen2024spacemantis}
Boyuan Chen, Zhuo Xu, Sean Kirmani, Brian Ichter, Danny Driess, Pete Florence,
  Dorsa Sadigh, Leonidas Guibas, and Fei Xia.
\newblock Spatialvlm: Endowing vision-language models with spatial reasoning
  capabilities.
\newblock \emph{arXiv preprint arXiv:2401.12168}, 2024{\natexlab{a}}.

\bibitem[Chen et~al.(2024{\natexlab{b}})Chen, Li, Dong, Zhang, Zang, Chen,
  Duan, Wang, Qiao, Lin, et~al.]{chen2024we}
Lin Chen, Jinsong Li, Xiaoyi Dong, Pan Zhang, Yuhang Zang, Zehui Chen, Haodong
  Duan, Jiaqi Wang, Yu Qiao, Dahua Lin, et~al.
\newblock Are we on the right way for evaluating large vision-language models?
\newblock \emph{Advances in Neural Information Processing Systems},
  37:\penalty0 27056--27087, 2024{\natexlab{b}}.

\bibitem[Chen et~al.(2024{\natexlab{c}})Chen, Wang, Cao, Liu, Gao, Cui, Zhu,
  Ye, Tian, Liu, et~al.]{chen2024expanding}
Zhe Chen, Weiyun Wang, Yue Cao, Yangzhou Liu, Zhangwei Gao, Erfei Cui, Jinguo
  Zhu, Shenglong Ye, Hao Tian, Zhaoyang Liu, et~al.
\newblock Expanding performance boundaries of open-source multimodal models
  with model, data, and test-time scaling.
\newblock \emph{arXiv preprint arXiv:2412.05271}, 2024{\natexlab{c}}.

\bibitem[Dai et~al.(2017)Dai, Chang, Savva, Halber, Funkhouser, and
  Nie{\ss}ner]{dai2017scannet}
Angela Dai, Angel~X Chang, Manolis Savva, Maciej Halber, Thomas Funkhouser, and
  Matthias Nie{\ss}ner.
\newblock Scannet: Richly-annotated 3d reconstructions of indoor scenes.
\newblock In \emph{Proceedings of the IEEE conference on computer vision and
  pattern recognition}, pages 5828--5839, 2017.

\bibitem[DeTone et~al.(2018)DeTone, Malisiewicz, and
  Rabinovich]{detone2018superpoint}
Daniel DeTone, Tomasz Malisiewicz, and Andrew Rabinovich.
\newblock Superpoint: Self-supervised interest point detection and description.
\newblock In \emph{Proceedings of the IEEE conference on computer vision and
  pattern recognition workshops}, pages 337--348, 2018.

\bibitem[Dongfang et~al.(2025)Dongfang, Zheng, Weng, Lyu, Paudel, Van~Gool,
  Yang, and Hu]{dongfang2025multimodal}
Zihao Dongfang, Xu Zheng, Ziqiao Weng, Yuanhuiyi Lyu, Danda~Pani Paudel, Luc
  Van~Gool, Kailun Yang, and Xuming Hu.
\newblock Are multimodal large language models ready for omnidirectional
  spatial reasoning?
\newblock \emph{arXiv preprint arXiv:2505.11907}, 2025.

\bibitem[Dusmanu et~al.(2019)Dusmanu, Rocco, Pajdla, Pollefeys, Sivic, Torii,
  and Sattler]{dusmanu2019d2}
Mihai Dusmanu, Ignacio Rocco, Tomas Pajdla, Marc Pollefeys, Josef Sivic,
  Akihiko Torii, and Torsten Sattler.
\newblock D2-net: A trainable cnn for joint detection and description of local
  features.
\newblock In \emph{CVPR 2019-IEEE Conference on Computer Vision and Pattern
  Recognition}, 2019.

\bibitem[Fischler and Bolles(1981)]{fischler1981random}
Martin~A Fischler and Robert~C Bolles.
\newblock Random sample consensus: a paradigm for model fitting with
  applications to image analysis and automated cartography.
\newblock \emph{Communications of the ACM}, 24\penalty0 (6):\penalty0 381--395,
  1981.

\bibitem[Gold and Rangarajan(2002)]{gold2002graduated}
Steven Gold and Anand Rangarajan.
\newblock A graduated assignment algorithm for graph matching.
\newblock \emph{IEEE Transactions on pattern analysis and machine
  intelligence}, 18\penalty0 (4):\penalty0 377--388, 2002.

\bibitem[Guo et~al.(2025)Guo, Yang, Zhang, Song, Wang, Zhu, Xu, Zhang, Ma, Bi,
  et~al.]{guo2025deepseek}
Daya Guo, Dejian Yang, Haowei Zhang, Junxiao Song, Peiyi Wang, Qihao Zhu,
  Runxin Xu, Ruoyu Zhang, Shirong Ma, Xiao Bi, et~al.
\newblock Deepseek-r1 incentivizes reasoning in llms through reinforcement
  learning.
\newblock \emph{Nature}, 645\penalty0 (8081):\penalty0 633--638, 2025.

\bibitem[Hartley(1994)]{hartley1994projective}
Richard~I Hartley.
\newblock Projective reconstruction and invariants from multiple images.
\newblock \emph{IEEE Transactions on Pattern Analysis and Machine
  Intelligence}, 16\penalty0 (10):\penalty0 1036--1041, 1994.

\bibitem[Hu et~al.(2024)Hu, Yin, Zhang, Cai, Long, Chen, Wang, Yu, Shen, and
  Shen]{hu2024metric3d}
Mu Hu, Wei Yin, Chi Zhang, Zhipeng Cai, Xiaoxiao Long, Hao Chen, Kaixuan Wang,
  Gang Yu, Chunhua Shen, and Shaojie Shen.
\newblock Metric3d v2: A versatile monocular geometric foundation model for
  zero-shot metric depth and surface normal estimation.
\newblock \emph{IEEE TPAMI}, 2024.

\bibitem[Huang et~al.(2025)Huang, Liu, Lin, Zhu, Zhao, Du, Li, Jia, Zhong,
  Chen, et~al.]{huang2025notvla}
Zheng Huang, Mingyu Liu, Xiaoyi Lin, Muzhi Zhu, Canyu Zhao, Zongze Du, Xiaoman
  Li, Yiduo Jia, Hao Zhong, Hao Chen, et~al.
\newblock Notvla: Narrowing of dense action trajectories for generalizable
  robot manipulation.
\newblock \emph{arXiv preprint arXiv:2510.03895}, 2025.

\bibitem[Ji et~al.(2025)Ji, Tan, Shi, Hao, Zhang, Zhang, Wang, Zhao, Mu, An,
  et~al.]{ji2025robobrain}
Yuheng Ji, Huajie Tan, Jiayu Shi, Xiaoshuai Hao, Yuan Zhang, Hengyuan Zhang,
  Pengwei Wang, Mengdi Zhao, Yao Mu, Pengju An, et~al.
\newblock Robobrain: A unified brain model for robotic manipulation from
  abstract to concrete.
\newblock \emph{arXiv preprint arXiv:2502.21257}, 2025.

\bibitem[Jia et~al.(2025)Jia, Qi, Zhang, Zhang, Yu, He, Wang, and
  Yi]{jia2025omnispatial}
Mengdi Jia, Zekun Qi, Shaochen Zhang, Wenyao Zhang, Xinqiang Yu, Jiawei He, He
  Wang, and Li Yi.
\newblock Omnispatial: Towards comprehensive spatial reasoning benchmark for
  vision language models.
\newblock \emph{arXiv preprint arXiv:2506.03135}, 2025.

\bibitem[Jin et~al.(2020)Jin, Mishkin, Mishchuk, Matas, Fua, Yi, and
  Trulls]{jin2020image}
Yuhe Jin, Dmytro Mishkin, Anastasiia Mishchuk, Jiri Matas, Pascal Fua,
  Kwang~Moo Yi, and Eduard Trulls.
\newblock Image matching across wide baselines: From paper to practice.
\newblock \emph{arXiv preprint arXiv:2003.01587}, 2020.

\bibitem[Kirillov et~al.(2023)Kirillov, Mintun, Ravi, Mao, Rolland, Gustafson,
  Xiao, Whitehead, Berg, Lo, et~al.]{kirillov2023segment}
Alexander Kirillov, Eric Mintun, Nikhila Ravi, Hanzi Mao, Chloe Rolland, Laura
  Gustafson, Tete Xiao, Spencer Whitehead, Alexander~C Berg, Wan-Yen Lo, et~al.
\newblock Segment anything.
\newblock In \emph{Proceedings of the IEEE/CVF international conference on
  computer vision}, pages 4015--4026, 2023.

\bibitem[Laurençon et~al.(2024)Laurençon, Marafioti, Sanh, and
  Tronchon]{laurencon2024idefics}
Hugo Laurençon, Andrés Marafioti, Victor Sanh, and Léo Tronchon.
\newblock Building and better understanding vision-language models: Insights
  and future directions.
\newblock In \emph{Workshop on Responsibly Building the Next Generation of
  Multimodal Foundational Models}, 2024.

\bibitem[Ling et~al.(2024)Ling, Sheng, Tu, Zhao, Xin, Wan, Yu, Guo, Yu, Lu,
  et~al.]{ling2024dl3dv}
Lu Ling, Yichen Sheng, Zhi Tu, Wentian Zhao, Cheng Xin, Kun Wan, Lantao Yu,
  Qianyu Guo, Zixun Yu, Yawen Lu, et~al.
\newblock Dl3dv-10k: A large-scale scene dataset for deep learning-based 3d
  vision.
\newblock In \emph{CVPR}, pages 22160--22169, 2024.

\bibitem[Liu et~al.(2024)Liu, Li, Li, and Lee]{liu2024llava15}
Haotian Liu, Chunyuan Li, Yuheng Li, and Yong~Jae Lee.
\newblock Improved baselines with visual instruction tuning.
\newblock In \emph{Proceedings of the IEEE/CVF Conference on Computer Vision
  and Pattern Recognition (CVPR)}, 2024.

\bibitem[Liu et~al.(2025)Liu, Tayal, Wang, Zarzar, Monnier, Tertikas, Duan,
  Toisoul, Zhang, Neverova, Vedaldi, Shapovalov, and Novotny]{liu24uco3d}
Xingchen Liu, Piyush Tayal, Jianyuan Wang, Jesus Zarzar, Tom Monnier,
  Konstantinos Tertikas, Jiali Duan, Antoine Toisoul, Jason~Y. Zhang, Natalia
  Neverova, Andrea Vedaldi, Roman Shapovalov, and David Novotny.
\newblock Uncommon objects in 3d.
\newblock In \emph{arXiv}, 2025.

\bibitem[Loshchilov and Hutter(2017)]{loshchilov2017decoupled}
Ilya Loshchilov and Frank Hutter.
\newblock Decoupled weight decay regularization.
\newblock \emph{arXiv preprint arXiv:1711.05101}, 2017.

\bibitem[Lowe(2004)]{lowe2004distinctive}
David~G Lowe.
\newblock Distinctive image features from scale-invariant keypoints.
\newblock \emph{International journal of computer vision}, 60\penalty0
  (2):\penalty0 91--110, 2004.

\bibitem[Ma et~al.(2025)Ma, Chen, Zhang, Chou, Chen, de~Melo, and
  Yuille]{ma20253dsrbench}
Wufei Ma, Haoyu Chen, Guofeng Zhang, Yu-Cheng Chou, Jieneng Chen, Celso de
  Melo, and Alan Yuille.
\newblock 3dsrbench: A comprehensive 3d spatial reasoning benchmark.
\newblock In \emph{Proceedings of the IEEE/CVF International Conference on
  Computer Vision}, pages 6924--6934, 2025.

\bibitem[Mishchuk et~al.(2017)Mishchuk, Mishkin, Radenovic, and
  Matas]{mishchuk2017working}
Anastasiia Mishchuk, Dmytro Mishkin, Filip Radenovic, and Jiri Matas.
\newblock Working hard to know your neighbor's margins: Local descriptor
  learning loss.
\newblock In \emph{Advances in neural information processing systems}, pages
  4826--4837, 2017.

\bibitem[{OpenAI}(2023)]{openai2023gpt4}
{OpenAI}.
\newblock Gpt-4 technical report.
\newblock \emph{CoRR, abs/2303.08774}, 2023.

\bibitem[OpenAI(2024)]{openai2024gpt4o}
OpenAI.
\newblock Hello gpt-4o.
\newblock \emph{OpenAI Blog}, 2024.
\newblock Accessed: November 22, 2024.

\bibitem[Pang et~al.(2025)Pang, Shao, Zhang, Tu, Liu, Zhou, Zhang, and
  Liu]{pang2025manivideo}
Youxin Pang, Ruizhi Shao, Jiajun Zhang, Hanzhang Tu, Yun Liu, Boyao Zhou,
  Hongwen Zhang, and Yebin Liu.
\newblock Manivideo: Generating hand-object manipulation video with dexterous
  and generalizable grasping.
\newblock In \emph{CVPR}, pages 12209--12219, 2025.

\bibitem[Pritchett and Zisserman(1998)]{pritchett1998wide}
Philip Pritchett and Andrew Zisserman.
\newblock Wide baseline stereo matching.
\newblock In \emph{ICCV}, pages 754--760. IEEE, 1998.

\bibitem[Ray et~al.(2024)Ray, Duan, Brown, Tan, Bashkirova, Hendrix, Ehsani,
  Kembhavi, Plummer, Krishna, Zeng, and Saenko]{ray2024sat}
Arijit Ray, Jiafei Duan, Ellis Brown, Reuben Tan, Dina Bashkirova, Rose
  Hendrix, Kiana Ehsani, Aniruddha Kembhavi, Bryan~A. Plummer, Ranjay Krishna,
  Kuo-Hao Zeng, and Kate Saenko.
\newblock Sat: Dynamic spatial aptitude training for multimodal language
  models.
\newblock \emph{arXiv preprint arXiv:2412.07755}, 2024.

\bibitem[Reizenstein et~al.(2021)Reizenstein, Shapovalov, Henzler, Sbordone,
  Labatut, and Novotny]{reizenstein2021common}
Jeremy Reizenstein, Roman Shapovalov, Philipp Henzler, Luca Sbordone, Patrick
  Labatut, and David Novotny.
\newblock Common objects in 3d: Large-scale learning and evaluation of
  real-life 3d category reconstruction.
\newblock In \emph{Proceedings of the IEEE/CVF international conference on
  computer vision}, pages 10901--10911, 2021.

\bibitem[Revaud et~al.(2019)Revaud, Weinzaepfel, De~Souza, Pion, Csurka, Cabon,
  and Humenberger]{revaud2019r2d2}
J{\'e}r{\^o}me Revaud, Philippe Weinzaepfel, C{\'e}sar De~Souza, No{\'e} Pion,
  Gabriela Csurka, Yohann Cabon, and Martin Humenberger.
\newblock R2d2: Repeatable and reliable detector and descriptor.
\newblock \emph{arXiv preprint arXiv:1906.06195}, 2019.

\bibitem[Rublee et~al.(2011)Rublee, Rabaud, Konolige, and
  Bradski]{rublee2011orb}
Ethan Rublee, Vincent Rabaud, Kurt Konolige, and Gary Bradski.
\newblock Orb: An efficient alternative to sift or surf.
\newblock In \emph{International conference on computer vision}, pages
  2564--2571. IEEE, 2011.

\bibitem[Schmid and Mohr(1995)]{schmid1995matching}
Cordelia Schmid and Roger Mohr.
\newblock \emph{Matching by local invariants}.
\newblock PhD thesis, INRIA, 1995.

\bibitem[Schonberger and Frahm(2016)]{schonberger2016structure}
Johannes~L Schonberger and Jan-Michael Frahm.
\newblock Structure-from-motion revisited.
\newblock In \emph{CVPR}, pages 4104--4113, 2016.

\bibitem[Song et~al.(2025)Song, Blukis, Tremblay, Tyree, Su, and
  Birchfield]{song2025robospatial}
Chan~Hee Song, Valts Blukis, Jonathan Tremblay, Stephen Tyree, Yu Su, and Stan
  Birchfield.
\newblock Robospatial: Teaching spatial understanding to 2d and 3d
  vision-language models for robotics.
\newblock In \emph{Proceedings of the Computer Vision and Pattern Recognition
  Conference}, pages 15768--15780, 2025.

\bibitem[Team et~al.(2023)Team, Anil, Borgeaud, Alayrac, Yu, Soricut,
  Schalkwyk, Dai, Hauth, Millican, et~al.]{gemini2023anil}
Gemini Team, Rohan Anil, Sebastian Borgeaud, Jean-Baptiste Alayrac, Jiahui Yu,
  Radu Soricut, Johan Schalkwyk, Andrew~M. Dai, Anja Hauth, Katie Millican,
  et~al.
\newblock Gemini: A family of highly capable multimodal models.
\newblock \emph{arXiv preprint arXiv:2312.11805}, 2023.

\bibitem[Tian et~al.(2019)Tian, Yu, Fan, Wu, Heijnen, and
  Balntas]{tian2019sosnet}
Yurun Tian, Xin Yu, Bin Fan, Fuchao Wu, Huub Heijnen, and Vassileios Balntas.
\newblock Sosnet: Second order similarity regularization for local descriptor
  learning.
\newblock In \emph{Proceedings of the IEEE/CVF conference on computer vision
  and pattern recognition}, pages 11494--11503, 2019.

\bibitem[Wang et~al.(2025{\natexlab{a}})Wang, Chen, Karaev, Vedaldi, Rupprecht,
  and Novotny]{wang2025vggt}
Jianyuan Wang, Minghao Chen, Nikita Karaev, Andrea Vedaldi, Christian
  Rupprecht, and David Novotny.
\newblock Vggt: Visual geometry grounded transformer.
\newblock In \emph{Proceedings of the Computer Vision and Pattern Recognition
  Conference}, pages 5294--5306, 2025{\natexlab{a}}.

\bibitem[Wang et~al.(2024)Wang, Bai, Tan, Wang, Fan, Bai, Chen, Liu, Wang, Ge,
  Fan, Dang, Du, Ren, Men, Liu, Zhou, Zhou, and Lin]{wang2024qwen2vl}
Peng Wang, Shuai Bai, Sinan Tan, Shijie Wang, Zhihao Fan, Jinze Bai, Keqin
  Chen, Xuejing Liu, Jialin Wang, Wenbin Ge, Yang Fan, Kai Dang, Mengfei Du,
  Xuancheng Ren, Rui Men, Dayiheng Liu, Chang Zhou, Jingren Zhou, and Junyang
  Lin.
\newblock Qwen2-vl: Enhancing vision-language model’s perception of the world
  at any resolution.
\newblock \emph{CoRR}, abs/2409.12191, 2024.

\bibitem[Wang et~al.(2025{\natexlab{b}})Wang, Xu, Dai, Xiang, Deng, Tong, and
  Yang]{wang2025moge}
Ruicheng Wang, Sicheng Xu, Cassie Dai, Jianfeng Xiang, Yu Deng, Xin Tong, and
  Jiaolong Yang.
\newblock Moge: Unlocking accurate monocular geometry estimation for
  open-domain images with optimal training supervision.
\newblock In \emph{CVPR}, pages 5261--5271, 2025{\natexlab{b}}.

\bibitem[Wang et~al.(2025{\natexlab{c}})Wang, Tan, Zhu, Yang, Yang, Wang,
  Kolobov, Gao, and Gong]{wang2025site}
Wenqi Wang, Reuben Tan, Pengyue Zhu, Jianwei Yang, Zhengyuan Yang, Lijuan Wang,
  Andrey Kolobov, Jianfeng Gao, and Boqing Gong.
\newblock Site: towards spatial intelligence thorough evaluation.
\newblock \emph{arXiv preprint arXiv:2505.05456}, 2025{\natexlab{c}}.

\bibitem[Wen et~al.(2025)Wen, Liu, Zheng, Ye, Wu, Wang, Xu, Liang, Li, Miao,
  et~al.]{wen2025reinforcement}
Xumeng Wen, Zihan Liu, Shun Zheng, Shengyu Ye, Zhirong Wu, Yang Wang, Zhijian
  Xu, Xiao Liang, Junjie Li, Ziming Miao, et~al.
\newblock Reinforcement learning with verifiable rewards implicitly
  incentivizes correct reasoning in base llms.
\newblock \emph{arXiv preprint arXiv:2506.14245}, 2025.

\bibitem[Wu and Xie(2024)]{wu2024v}
Penghao Wu and Saining Xie.
\newblock V?: Guided visual search as a core mechanism in multimodal llms.
\newblock In \emph{Proceedings of the IEEE/CVF Conference on Computer Vision
  and Pattern Recognition}, pages 13084--13094, 2024.

\bibitem[xAI(2024)]{realworldqa2024}
xAI.
\newblock Realworldqa: Real-world visual question answering benchmark.
\newblock \url{https://x.ai/news/grok-1.5v}, 2024.

\bibitem[Xu et~al.(2025)Xu, Wang, Tang, Chen, Wang, Chu, Lin, Feiszli, and
  Liang]{xu2025multi}
Runsen Xu, Weiyao Wang, Hao Tang, Xingyu Chen, Xiaodong Wang, Fu-Jen Chu, Dahua
  Lin, Matt Feiszli, and Kevin~J Liang.
\newblock Multi-spatialmllm: Multi-frame spatial understanding with multi-modal
  large language models.
\newblock \emph{arXiv preprint arXiv:2505.17015}, 2025.

\bibitem[Yang et~al.(2025)Yang, Yang, Gupta, Han, Fei-Fei, and
  Xie]{yang2025thinking}
Jihan Yang, Shusheng Yang, Anjali~W Gupta, Rilyn Han, Li Fei-Fei, and Saining
  Xie.
\newblock Thinking in space: How multimodal large language models see,
  remember, and recall spaces.
\newblock In \emph{Proceedings of the Computer Vision and Pattern Recognition
  Conference}, pages 10632--10643, 2025.

\bibitem[Yeh et~al.(2025)Yeh, Wang, Tong, Cheng, Wang, Chu, Zhai, Chen, Gao,
  and Ma]{yeh2025seeing}
Chun-Hsiao Yeh, Chenyu Wang, Shengbang Tong, Ta-Ying Cheng, Ruoyu Wang, Tianzhe
  Chu, Yuexiang Zhai, Yubei Chen, Shenghua Gao, and Yi Ma.
\newblock Seeing from another perspective: Evaluating multi-view understanding
  in mllms.
\newblock \emph{arXiv preprint arXiv:2504.15280}, 2025.

\bibitem[Yin et~al.(2025)Yin, Wang, Zhang, Zhang, Wang, Wang, Zhang,
  Chandrasegaran, Liu, Krishna, Xie, Li, Wu, and Fei-Fei]{yin2025mindcube}
Baiqiao Yin, Qineng Wang, Pingyue Zhang, Jianshu Zhang, Kangrui Wang, Zihan
  Wang, Jieyu Zhang, Keshigeyan Chandrasegaran, Han Liu, Ranjay Krishna,
  Saining Xie, Manling Li, Jiajun Wu, and Li Fei-Fei.
\newblock Mindcube: Spatial mental modeling from limited views.
\newblock \emph{arXiv preprint arXiv:2506.21458}, 2025.

\bibitem[Zhang et~al.(2024{\natexlab{a}})Zhang, Li, Zhang, Pu, Cahyono, Hu,
  Liu, Zhang, Yang, Li, and Liu]{lmms_eval}
Kaichen Zhang, Bo Li, Peiyuan Zhang, Fanyi Pu, Joshua~Adrian Cahyono, Kairui
  Hu, Shuai Liu, Yuanhan Zhang, Jingkang Yang, Chunyuan Li, and Ziwei Liu.
\newblock Lmms-eval: Reality check on the evaluation of large multimodal
  models, 2024{\natexlab{a}}.

\bibitem[Zhang et~al.(2024{\natexlab{b}})Zhang, Zhang, Li, Zeng, Yang, Zhang,
  Wang, Tan, Li, and Liu]{zhang2024longva}
Peiyuan Zhang, Kaichen Zhang, Bo Li, Guangtao Zeng, Jingkang Yang, Yuanhan
  Zhang, Ziyue Wang, Haoran Tan, Chunyuan Li, and Ziwei Liu.
\newblock Long context transfer from language to vision.
\newblock \emph{arXiv preprint arXiv:2406.16852}, 2024{\natexlab{b}}.

\bibitem[Zhang et~al.(2024{\natexlab{c}})Zhang, Zhang, Tian, Fu, Zhang, Wu, Li,
  Wang, Wen, Zhang, et~al.]{zhang2024mme}
Yi-Fan Zhang, Huanyu Zhang, Haochen Tian, Chaoyou Fu, Shuangqing Zhang, Junfei
  Wu, Feng Li, Kun Wang, Qingsong Wen, Zhang Zhang, et~al.
\newblock Mme-realworld: Could your multimodal llm challenge high-resolution
  real-world scenarios that are difficult for humans?
\newblock \emph{arXiv preprint arXiv:2408.13257}, 2024{\natexlab{c}}.

\bibitem[Zhong et~al.(2025)Zhong, Zhu, Du, Huang, Zhao, Liu, Wang, Chen, and
  Shen]{zhong2025omni}
Hao Zhong, Muzhi Zhu, Zongze Du, Zheng Huang, Canyu Zhao, Mingyu Liu, Wen Wang,
  Hao Chen, and Chunhua Shen.
\newblock Omni-r1: Reinforcement learning for omnimodal reasoning via
  two-system collaboration.
\newblock \emph{arXiv preprint arXiv:2505.20256}, 2025.

\bibitem[Zhou et~al.(2025)Zhou, An, Chi, Han, Rong, Zhang, Wang, Wang, Huang,
  Sheng, et~al.]{zhou2025roborefer}
Enshen Zhou, Jingkun An, Cheng Chi, Yi Han, Shanyu Rong, Chi Zhang, Pengwei
  Wang, Zhongyuan Wang, Tiejun Huang, Lu Sheng, et~al.
\newblock Roborefer: Towards spatial referring with reasoning in
  vision-language models for robotics.
\newblock \emph{arXiv preprint arXiv:2506.04308}, 2025.

\bibitem[Zhou et~al.(2018)Zhou, Tucker, Flynn, Fyffe, and
  Snavely]{zhou2018stereo}
Tinghui Zhou, Richard Tucker, John Flynn, Graham Fyffe, and Noah Snavely.
\newblock Stereo magnification: learning view synthesis using multiplane
  images.
\newblock \emph{ACM Transactions on Graphics (TOG)}, 37\penalty0 (4):\penalty0
  1--12, 2018.

\bibitem[Zhu et~al.(2025{\natexlab{a}})Zhu, Wang, Chen, Liu, Ye, Gu, Duan,
  Tian, Su, Shao, et~al.]{zhu2025internvl3}
Jinguo Zhu, Weiyun Wang, Zhe Chen, Zhaoyang Liu, Shenglong Ye, Lixin Gu, Yuchen
  Duan, Hao Tian, Weijie Su, Jie Shao, et~al.
\newblock Internvl3: Exploring advanced training and test-time recipes for
  open-source multimodal models.
\newblock \emph{CoRR, abs/2504.10479}, 2025{\natexlab{a}}.

\bibitem[Zhu et~al.(2025{\natexlab{b}})Zhu, Tian, Chen, Zhou, Guo, Liu, Yang,
  and Shen]{zhu2025segagent}
Muzhi Zhu, Yuzhuo Tian, Hao Chen, Chunluan Zhou, Qingpei Guo, Yang Liu, Ming
  Yang, and Chunhua Shen.
\newblock Segagent: Exploring pixel understanding capabilities in mllms by
  imitating human annotator trajectories.
\newblock In \emph{Proceedings of the Computer Vision and Pattern Recognition
  Conference}, pages 3686--3696, 2025{\natexlab{b}}.

\bibitem[Zhu et~al.(2025{\natexlab{c}})Zhu, Zhong, Zhao, Du, Huang, Liu, Chen,
  Zou, Chen, Yang, et~al.]{zhu2025active}
Muzhi Zhu, Hao Zhong, Canyu Zhao, Zongze Du, Zheng Huang, Mingyu Liu, Hao Chen,
  Cheng Zou, Jingdong Chen, Ming Yang, et~al.
\newblock Active-o3: Empowering multimodal large language models with active
  perception via grpo.
\newblock \emph{arXiv preprint arXiv:2505.21457}, 2025{\natexlab{c}}.

\end{thebibliography}
\end{document}